\documentclass[sigconf]{acmart}

\renewcommand\footnotetextcopyrightpermission[1]{} % removes footnote with conference information in first column
\usepackage{booktabs} % For formal tables
\usepackage{amsmath}
\usepackage{amsfonts}
\usepackage{graphicx}
\usepackage{array,multirow}
\usepackage{algorithmicx}
\usepackage{algorithm}
\usepackage[noend]{algpseudocode}
\usepackage{hyperref}

\makeatletter
\def\BState{\State\hskip-\ALG@thistlm}
\makeatother

\newcommand*\rot{\rotatebox{90}}

\makeatletter
\def\@copyrightspace{\relax}
\makeatother

\begin{document}
\title[Novelty Search for Deep RL Network Weights By Edit Metric Distance]{Novelty Search for Deep Reinforcement Learning Policy Network Weights by Action Sequence Edit Metric Distance}

%\titlenote{Produces the permission block, and
%  copyright information}
%\subtitle{Subtitle}
%\subtitlenote{The full version of the author's guide is available as
%  \texttt{acmart.pdf} document}

%%% The submitted version for review should be ANONYMOUS
\author{Ethan C. Jackson}
\affiliation{%
  \institution{The University of Western Ontario\\ Vector Institute}
}
\email{jackson.ethan.c@gmail.com}

\author{Mark Daley}
\affiliation{%
	\institution{The University of Western Ontario\\ Vector Institute}
}
\email{mdaley2@uwo.ca}

\renewcommand{\shortauthors}{Ethan C. Jackson and Mark Daley}

\begin{abstract}
Reinforcement learning (RL) problems often feature deceptive local optima, and learning methods that optimize purely for reward signal often fail to learn strategies for overcoming them \cite{lehman2011abandoning}. Deep neuroevolution and novelty search have been proposed as effective alternatives to gradient-based methods for learning RL policies directly from pixels. In this paper, we introduce and evaluate the use of novelty search over agent action sequences by string edit metric distance as a means for promoting innovation. We also introduce a method for stagnation detection and population resampling inspired by recent developments in the RL community \cite{savinov2018episodic}, \cite{goexplore} that uses the same mechanisms as novelty search to promote and develop innovative policies. Our methods extend a state-of-the-art method for deep neuroevolution using a simple-yet-effective genetic algorithm (GA) designed to efficiently learn deep RL policy network weights \cite{such2017deep}. Experiments using four games from the Atari 2600 benchmark were conducted. Results provide further evidence that GAs are competitive with gradient-based algorithms for deep RL. Results also demonstrate that novelty search over action sequences is an effective source of selection pressure that can be integrated into existing evolutionary algorithms for deep RL. 
\end{abstract}

\maketitle

\def\etal.{et\penalty50\ al.}

\section{Introduction}

%Artificial intelligence (AI) and neural networks have a long history of incorporating memory into models. Recurrent neural networks (RNNs) and Long Short-Term Memory (LSTM) networks are canonical examples of models capable of storing and recalling previous inputs or network states. In recent work, various approaches have been developed to use environmental or observational memory to help overcome learning problems in RL. In particular, the problems of \textit{deceptive local optima} or \textit{traps} and \textit{sparse reward} often appear in RL benchmarks and applications. 

%Many existing strategies for avoiding deceptive local optima or dealing with sparse reward work by rewarding or promoting agents that explore their environments differently than others. Exploration is often quantified using information contained in environment observations. For example, agents that reach new locations, or that reach the same location using an alternate route, can be rewarded for their novel behaviour. 

%To further develop measures of novelty or curiosity that could be widely useful in RL, we propose that --- in addition to environmentally-derived information --- an agent's behaviour could be more directly leveraged during learning. Specifically, the sequence of actions performed by an agent in response to observations over time could be used to characterize its behaviour. Then, a variety of metrics could be used to compute behavioural novelty between agents based solely on their actions. 

Reinforcement learning (RL) \cite{sutton1998introduction} problems often feature \textit{deceptive local optima} that impose difficult challenges to many learning algorithms. Algorithms that optimize strictly for reward often produce \textit{degenerate policies} that cause agents to under-explore their environments or under-develop strategies for increasing reward. Deceptive local optima have proved to be equally challenging for both gradient-based RL algorithms, including DQN \cite{mnih2015human}, and gradient-free algorithms including genetic algorithms (GAs) \cite{such2017deep}. 

Deceptive local optima in reinforcement learning have long been studied by the evolutionary algorithms community --- with concepts including \textit{novelty search} being introduced in response \cite{lehman2011abandoning}. The deep RL community has responded with similar ideas and tools, but in purely gradient-based learning frameworks. A good example is given by recent work from Google Brain and DeepMind that promotes \textit{episodic curiosity} in deep RL benchmarks \cite{savinov2018episodic}. These methods were both designed to address deceptive local optima by substituting or supplementing reward signal with some measure of behavioural novelty. In practice, an agent's behaviour has usually been defined in terms of its environment.  Behaviour is often quantified using information contained in environment observations. For example, agents that reach new locations \cite{such2017deep}, or that reach the same location using an alternate route\cite{savinov2018episodic}, can be rewarded for their novel behaviour. 

In this paper, we investigate whether agent behaviour can be quantified more generally and leveraged more directly. We investigate the following question: ``\textit{Can the history of actions performed by agents be used to promote innovative behaviour in benchmark RL problems?}' Towards answering this, we implemented two novel methods for incorporating behavioural history in an evolutionary algorithm designed to effectively train deep RL networks. The base algorithm is an approximate replication of Such \etal.'s genetic algorithm (GA) for learning DQN network \cite{mnih2015human} weights. This is a very simple yet effective gradient-free approach for learning DQN policies that are competitive with those produced by Deep Q-learning \cite{such2017deep}.

%To ensure comparability of results, we used the same architecture in all experiments and reimplemented the same GA as a control (Base GA). 

%Using four games from the Atari 2600 benchmark [] --- \textsc{Assault}, \textsc{Asteroids}, \textsc{MsPacman}, and \textsc{Space Invaders} ---

Both methods are GA extensions based on Lehman and Stanley's \textit{novelty search} \cite{lehman2011abandoning} --- an evolutionary algorithm designed to avoid deceptive local optima by defining selection pressure in terms of behaviour instead of conventional optimization criteria such as reward signal. Novelty search has been shown to be an effective tool for promoting innovation in RL \cite{such2017deep}. In this paper, we introduce the use of Levenshtein distance \cite{levenshtein1966binary} --- a form of \textit{string edit metric distance} --- as the behavioural distance function in a novelty search for deep RL network weights.   

The first method (Method I) is an implementation of novelty search in which, during training, the reward signal is completely substituted by a novelty score based on the Levenshtein distance between sequences of game actions. In a novelty search, \textit{behaviour characteristics} are stored in an \textit{archive} for a randomly-selected subset of individuals in each generation. We define the behaviour characteristic as the sequence of actions performed by an agent during the training episode. Selection pressure is then determined by computing the \textit{behavioural distance} between individuals in the current population and those in the archive --- which we define as Levenshtein distance.

The second method (Method II) is not a novelty search, but rather a modification to the Base GA that incorporates elements of novelty search to avoid population convergence to locally-optimal behaviours. The modified algorithm detects slowing learning progress as measured using game scores in validation episodes. When validation scores are non-increasing for a fixed number of episodes, the population is regenerated by sampling the archive for individuals whose behaviours were most novel compared to the current population --- a concept related to restarting and recentering in evolutionary algorithms \cite{hughes2013recentering}.

Using two sets of experiments, we evaluated each method's effectiveness for learning RL policies for four Atari 2600 games, namely \textsc{Assault}, \textsc{Asteroids}, \textsc{MsPacman}, and \textsc{Space Invaders}. We found that while Method I is less effective than the Base GA for learning high-scoring policies, it returns policies that are behaviourally distinct. For example, we observed greater uses of obstacles or greater agent lifespans in some games. Method II was more effective than Method I for learning high-scoring policies. In two out of four games, it produced better-scoring policies than the Base GA, and in one out of four, it produced better-scoring policies than the original DQN learning method. 

Importantly, and in contrast to previous uses of novelty search for deep RL, the behaviour characteristic and behavioural distance function used here \textit{do not require environment-specific knowledge}. While such a requirement is not inherently a hindrance, it is convenient to have tools that work in more general contexts. Compared to related methods that use memories of observations (usually environment observations) to return to previous states \cite{goexplore} or to re-experience or re-visit under-explored areas \cite{savinov2018episodic}, archives of action sequences are relatively compact, easy to store, and efficient to compare. As such, the methods presented in this paper can either be used as stand-alone frameworks, or as extensions to existing methods that use environmental memory to improve learning.

In the next section, we give an overview of the Base GA and architecture, the Atari benchmark problem, and our experimental setup. In Section \ref{section:novelty} we provide a full definition of novelty search and details of our implementation based on action sequences and Levenshtein distance (Method I). In Section \ref{section:resampling} we provide further details for Method II. Section \ref{section:experiments} describes experiments and results, and is followed by discussion in Section \ref{section:discussion}.

\section{Highly-Scalable Genetic Algorithms for Deep Reinforcement Learning}\label{section:GA}

The conventional objective in RL is to produce an optimal \textit{policy} --- a function that maps states to actions such that \textit{reward}, or gain in reward, is optimized. The methods introduced in this paper are extensions of a replicated state-of-the-art GA for learning deep RL policy network weights introduced by Such \etal. in \cite{such2017deep}. 

\subsection{DQN Architecture and Preprocessing}

A RL policy network is an instance of a neural network that implements a RL policy. For comparability to related work, we used the DQN neural network architecture \cite{mnih2015human} in all experiments. This network consists of three convolutional layers with 32, 64, and 64 filters, respectively, followed by one dense layer with 512 units. The convolutional filter sizes are $8 \times 8$, $4 \times 4$, and $3 \times 3$, respectively. The strides are 4, 2, and 1, respectively.  All network weights are initialized using Glorot normal initialization. All network layer outputs use rectified linear unit (ReLU) activation. All game observations (frames) are downsampled to $84 \times 84 \times 4$ arrays. The third dimension reflects separate intensity channels for red, green, blue, and luminosity. Consecutive game observations are summed to rectify sprite flickering. 

\subsection{Seed-Based Genetic Algorithm}

Perhaps surprisingly, very simple genetic algorithms have been shown to be competitive with Deep Q-learning for learning DQN architecture parameterizations \cite{such2017deep}. In their paper, Such \etal. introduced an efficient seed-based encoding that enables very large network parameterizations to be indirectly encoded by a list of deterministic pseudo-random number generator (PRNG) seeds. This, in contrast to a direct encoding, scales with the number of evolutionary generations (typically thousands) rather than the number of network connections (typically millions or more). This encoding enables GAs to work at unprecedented scales for tuning neural network weights. It thus enables, more generally, a new wave of exploration for evolutionary algorithms and deep learning. 

For the present work, we implemented a GA and encoding approximately as described in \cite{such2017deep} using Keras \cite{Keras}, a high-level interface for Tensorflow \cite{abadi2016tensorflow}, and NumPy \cite{oliphant2006guide}. An individual in the GA's population is encoded by a list of seeds for both Keras' and NumPy's deterministic PRNGs. The first seed is used to initialize network weights. Subsequent seeds are used to produce additive mutation noise. A constant scaling factor (mutation power) is used to scale down the intensity of noise added per generation. 

A network parameterization is thus defined by:
\begin{align}
\Theta^n &= \Theta^{n-1} + \sigma \epsilon(\tau_n)\\
\Theta^0 &= \phi(\tau_0)
\end{align}

\noindent where $\Theta^n$ denotes network weights at generation $n$, $\tau$ denotes the encoding of $\Theta^n$ as a list of seeds, $\phi$ denotes a seeded, deterministic initialization function, $\epsilon(\tau_n) \sim \mathcal{N}(0,1)$ denotes a seeded, deterministic, normally-distributed PRNG seeded with $\tau_n$\, and $\sigma$ denotes a constant scaling factor (mutation power). 

As in its introductory paper, the GA does not implement crossover, and mutation simply appends a randomly-generated seed to an individual's list $\tau$. The GA performs \textit{truncated selection} --- a process whereby the top $T$ individuals are selected as reproduction candidates (parents) for the next generation. From these $T$ parents, the next generation's population is uniformly, randomly sampled with replacement, and mutated. 

The GA also implements a form of \textit{elitism} --- a commonly used tactic to ensure that the best performing individual is preserved in the next generation without mutation. A separate set of \textit{validation episodes} is used to help determine the elite individual during training. This has the effect of adding secondary selection pressure for generalizability and helps to reduce overfitting. More details are given in Section \ref{section:experiments}.

It is important to note that this encoding imposes network reconstruction costs that would not be needed using a direct encoding. The compact representation, though, enables a high degree of scalability that would not be practical using a direct encoding. Algorithm descriptions and source code for the Base GA, Method I, and Method II are provided in the Appendix and \hyperref{https://github.com/ethancjackson/NoveltySearchLevenshtein}{}{}{Digital Appendix}\footnote{https://github.com/ethancjackson/NoveltySearchLevenshtein}, respectively. For further details on the Base GA, refer to \cite{such2017deep}.

\subsection{Atari 2600 Benchmark}

The Atari 2600 Benchmark is provided as part of OpenAIGym \cite{brockman2016openai} --- an open-source platform for experimenting with a variety of reinforcement learning problems. Work by Mnih \etal. \cite{mnih2013playing} introduced a novel method and architecture for learning to play games directly from pixels --- a challenge that remains difficult \cite{hessel2018rainbow}. Though many enhancements and extensions have been developed for DQN, no single learning method has emerged as being generally dominant \cite{hessel2018rainbow}, \cite{goexplore}. 

The games included in the Atari 2600 benchmark provide a diverse set of control problems. In particular, the games vary greatly in both gameplay and logic. In \textsc{MsPacman}, for example, part of the challenge comes from the fact that the rules for success change once \textsc{MsPacman} consumes a pill. To achieve a high score, the player or agent must shift strategies from escape to pursuit. This is quite different from \textsc{Breakout} for example --- a game in which the optimal paddle position can be computed as a function of consecutive ball observations. The variety of problems provided by this benchmark makes it an interesting set to study.

Before designing experiments, it is important to ask whether the chosen methods are plausibly capable of learning high quality policies. In games like \textsc{MsPacman}, is it reasonable to expect that a strictly feed-forward network architecture like DQN should be capable of producing high-quality policies? Though we do not investigate this question in the experiments presented in this paper, we comment on it in Section \ref{section:discussion}.

\subsection{Experimental Setup}

The Base GA and encoding for our experiments is an approximate replication of the GA and encoding introduced by Such \etal. in \cite{such2017deep}. All code was written in Python and uses Keras and TensorFlow for network evaluation. All experiments were run on a CPU-only 32-core Microsoft Azure cloud-based virtual machine (Standard F32s\_v2). The code is scalable to any number of threads and could be adapted to run on a distributed system. A single run of a Method II experiment (see Table \ref{table:method2params}) required roughly 120 wall-hours of compute time using this system.

\section{Novelty Search Over Action Sequences}\label{section:novelty}

\begin{figure}
	\centering
	\includegraphics[width=0.7\linewidth]{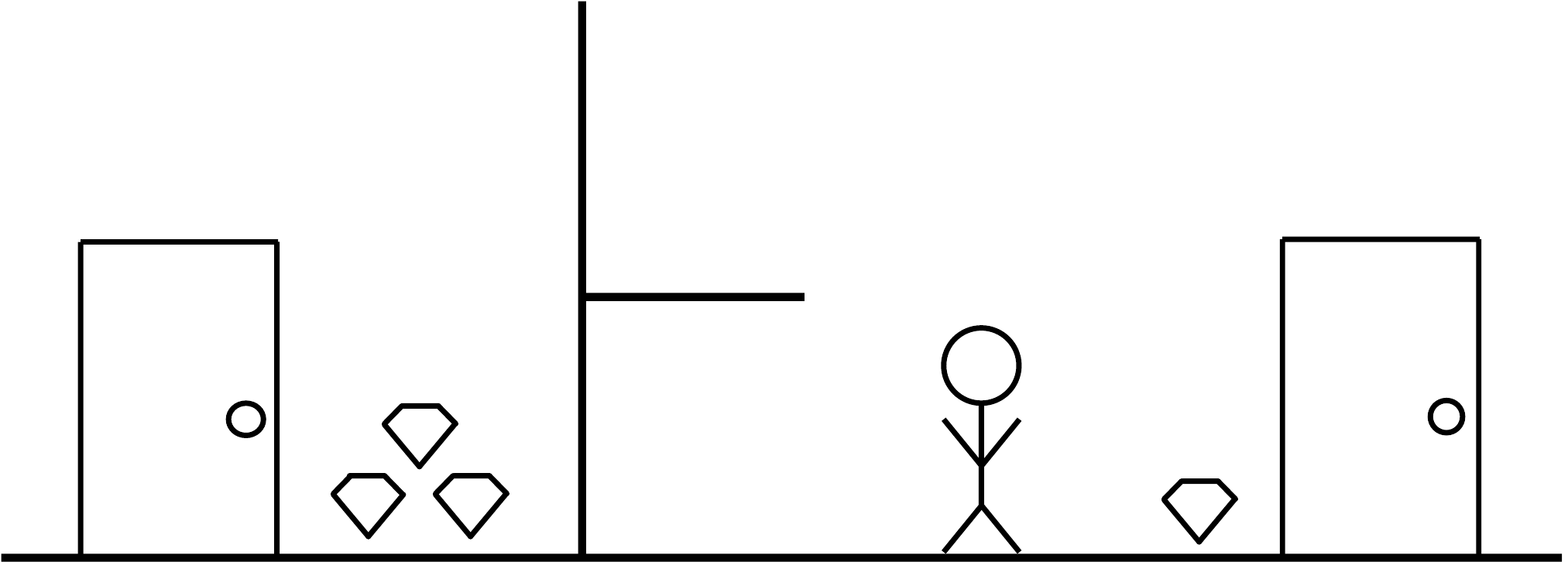}
	\caption{Example of a simple game stage with a deceptive local optimum. Assuming the goal is for the player to earn points by collecting as many diamonds as possible before using a door to exit the stage, a globally suboptimal policy may never learn to scale the wall to the player's left and collect three additional diamonds.}
	\label{fig:localoptima}
\end{figure}

Reinforcement learning problems often feature deceptive local optima or sparse reward signals. For example, consider a simple platform game in which the player navigates the environment to collect rewards.  Environmental obstacles, such as walls and stacked platforms, increase gameplay complexity and introduce latent optimization criteria. A simple example of such a game is visualized by Figure \ref{fig:localoptima}.

To overcome such challenges, agents may need to develop behavioural or strategic innovations that are not exhibited by any agent in the initial population. While it is possible for innovations to appear strictly as a result of mutations using the Base GA, these innovations are only promoted to the next generation if they immediately yield a positive return in terms of reward signal. Introduced by Lehman and Stanley in \cite{lehman2011abandoning}, novelty search addresses environmental challenges in RL by redefining optimization criteria in terms of behavioural innovation. In the context of evolutionary algorithms, including GAs, a pure novelty search defines fitness in terms of novelty rather than reward. Novelty search requires the following additional components over a typical genetic algorithm: 1) a behaviour characteristic, 2) a behavioural distance function, and 3) an archive of stored behaviour characteristics. 

\subsection{Behaviour Characteristic}

The behaviour characteristic of a policy $\pi$, denoted by $BC(\pi)$, is a description of some behavioural aspect of $\pi$ with respect to its environment. For example, the behaviour characteristic of a policy for controlling an agent in a 2D navigation environment could be the coordinate pair of the agent's location at the end of an episode. The behavioural distance between two behaviour characteristics is the output of a suitable distance metric function $d$ applied to two behaviour characteristics $BC(\pi_i)$ and $BC(\pi_j)$. For example, assuming that $BC(\pi)$ maps a policy $\pi$ to the final resting coordinates of an agent in 2D space, the behavioural distance function $d$ could be Euclidean distance in $\mathbb{R}^2$. Continuing with this example, an archive would consist of a randomly-selected subset of final resting coordinates reached by agents throughout training.

In previous work, both behaviour characteristics and behavioural distance functions were assumed to be domain-specific: they would not usually generalize to other environments. In this paper, we introduce a generalized formulation of novelty that applies to any game in the Atari 2600 benchmark, and that generalizes to many more control problems. 

We define the behaviour characteristic of a policy to be the sequence of discrete actions performed by an agent in response to consecutive environment observations. These action sequences are encoded as strings of length $F$, where $F$ is the maximum number of frames available during training. Characters are either elements of a game's \textit{action space} (distinct symbols that encode a button press) or the character $x$, which is reserved to encode a \textit{death action} or non-consumed frame. 

\subsection{Behavioural Distance Function}

We define the behavioural distance function as an approximation of the Levenshtein distance \cite{levenshtein1966binary} between action sequences encoded by strings. Note that other string edit distance metrics, such as Hamming distance \cite{hamming1950error}, or distributional distance metrics, such as Kullback-Leibler divergence \cite{kullback1951information} could also be used as behavioural distance functions. We chose to base our behavioural distance function on Levenshtein distance because it captures temporal relationships between action sequences that the other metrics do not. 

For example, two action sequences encoded by \textit{x12345} and \textit{12345x} are much closer in Levenshtein space (two edits: one deletion and one insertion) than in Hamming space (six edits: one substitution at each position). The Kullback-Leibler divergence between the distribution of actions in these two strings is zero since each action occurs exactly once, thus failing to discriminate the two policies by their statistics.

The additional descriptive power of Levenshtein distance comes with higher computational costs. The time complexity of computing the Levenshtein distance between two strings of length $n$ is $O(n^2)$. For large enough $n$, Levenshtein distance computations will impose a bottleneck on learning -- a problem we encountered in preliminary experiments.

To remedy this, we simply restrict the size of $n$ by splitting action sequences into fixed-length segments and compute the cumulative Levenshtein distance between corresponding segments. All experiments reported in this paper use $n=500$ for computing segmented Levenshtein distance. While some information is lost using this approach, the practical reduction in runtime necessitates the choice. The behavioural distance function $d$  which computes segmented Levenshtein distance is defined by Equation \ref{bdist}:

\begin{align}\label{bdist}
d(A,B) = \sum_{s=0}^{S-1} L(A_{sn \cdots sn+n-1}, B_{sn \cdots sn+n-1})
\end{align}

\noindent where $A$ and $B$ are two action sequences encoded by strings, $S$ is the number of segments, $n$ is the length of each segment, and $L$ computes the Levenshtein distance between two strings. The number of segments $n$ is determined by computing $\lceil F / s \rceil$, where $F$ is the number of characters in $A$ and $B$, equal to the maximum number of frames available during training. In experiments, $L$ is computed using the Python package python-Levenshtein \cite{python-levenshtein}.

\subsection{Hybrid Algorithm}

In a pure novelty search, fitness in the GA would be defined entirely by novelty scores. The experiments reported in this paper for Method I use a hybrid algorithm in which, like for a pure novelty search, selection pressure during training is solely determined using novelty scores. To identify the generation elite, however, we use the validation game score instead of a novelty score. This is due to the episodic nature of our chosen behaviour characteristic. Action sequences archived during training are specific to the training episode. To be consistent with other experiments using novelty search, we avoided the introduction of validation-specific archives for additional episodes. And so while novelty is the dominant component of selection pressure, we make this distinction clear to differentiate it from a pure novelty search. Experiments using Method I are discussed in Section \ref{sec:method1}.

\section{Novelty-Based Population Resampling in Genetic Algorithms}\label{section:resampling}

Reward sparsity is highly variable between RL problems. The Atari 2600 game \textsc{Montezuma's Revenge}, for example, is a complex platform game that requires significant exploration, puzzle-solving, and other strategies to complete. Until very recently, it has proved challenging to develop high-performing policies for this game without human-generated playthrough examples. A new method called \textit{Go-Explore} was recently introduced as the state-of-the-art for producing \textsc{Montezuma's Revenge} policies \cite{goexplore}. Though it is not based on evolutionary algorithms or the DQN architecture, Go-Explore borrows ideas from novelty search -- namely the use of an archive to store and recall states over the course of policy search. 

Motivated by this result, we designed Method II as an extension to the Base GA that adds features inspired by Go-Explore. In particular, we designed experiments to test whether an archive of action-sequences recorded throughout evolution could be effectively used for promoting innovation. Over the course of evolution, a randomly selected subset of individuals together with their action sequences are archived. This archive gradually collects individuals that could potentially lead to better policies than those that were selected for reproduction. Since the Base GA's selection pressure is based entirely on game score, it is still susceptible to converge around locally optimal policies and to discard innovations that do not yield immediate returns. 

Since novelty scores are not computed to determine primary selection pressure, Method II is not a novelty search. Instead, novelty scores are only computed when the algorithm detects that policy generalizability has stagnated over some number of generations. In such cases, the algorithm generates a new population by sampling the archive for individuals whose behaviour characteristics are most distant from the current population. These sampled individuals are used as parents for the next generation and the GA proceeds otherwise identically as the Base GA. 

As expected given DQN's prior ineffectiveness for learning \textsc{Montezuma's Revenge}, both Methods I and II were also unsuccessful in preliminary experiments. As a result we excluded it from main experiments, which are discussed in the next Section.

\section{Experiments}\label{section:experiments}

All experiments use the same four games: \textsc{Assault}, \textsc{Asteroids}, \textsc{MsPacman}, and \textsc{Space Invaders}. These games were chosen because they each feature gameplay that falls into one of two categories: games with one- or two-dimensional navigation. \textsc{Assault} and \textsc{Space Invaders} both allow the player or agent to move an avatar across a one-dimensional axis at the bottom of the game screen, while \textsc{Asteroids} and \textsc{MsPacman} allow a much greater range of exploration. Experiments using these four games also provide new results for the Base GA's effectiveness for learning to play Atari using the DQN architecture. 

In all experiments, we provide a baseline result using our replication of the GA described by Such \etal. in \cite{such2017deep}. The purpose of this baseline is to provide a replicated benchmark for using GAs to learn DQN architecture weights. While we acknowledge that many modified versions of the DQN architecture have been developed \cite{hessel2018rainbow}, we use the original architecture to ensure comparability to a wide variety of existing results, thereby controlling for differences between algorithms rather than architectures. Video comparisons of the Base GA and Methods I and II are included in the Digital Appendix.
\begin{table}
	\begin{center}
		\begin{tabular}{lrr}
			\hline
			Hyperparameter & Method I &   Method II \\ 
			\hline 
			Population Size (N) & 100 + 1& 1,000 + 1 \\ 
			
			Generations & 500 & 1000 \\ 
			
			Truncation Size (T) & 20 & 20 \\ 
			
			Mutation Power ($\sigma$) & 0.002 & 0.002 \\ 
			
			Archive Probability & 0.1 & 0.01 \\ 
			
			Max Frames Per Episode (F) & 20,000 & 20,000 \\ 
			
			Training Episodes & 1 & 1 \\ 
			
			Validation Episodes  & 5 & 30 \\ 
			
			Improvement Generations (\emph{IG}) &  & 10 \\ 
			\hline 
		\end{tabular} 
	\end{center}
	
	\caption{Hyperparameters for Method I and Method II experiments. Note that the Improvement Generations hyperparameter is only used in Method II experiments, and that baseline results do not use archiving. Population sizes are incremented to account for elites.}\label{table:method2params}
\end{table}

\subsection{Method I}\label{sec:method1}

Method I was designed to test the merits of using novelty search over agent action sequences in the Atari 2600 benchmark. This method substitutes reward signal with a measure of behavioural novelty as the selection pressure in an evolutionary search for DQN architecture weights. For comparability with existing gradient-based \cite{mnih2015human} and gradient-free \cite{such2017deep} methods, we evaluated Method I against the Base GA. Due to compute time constraints, these experiments were run at a smaller scale than for Method II. Hyperparameters are summarized by Table \ref{table:method2params}. 

\begin{figure*}
	\begin{tabular}{ccccc}
		\multicolumn{5}{c}{Method I Learning Progress}\\
		\hline
		& \textsc{Assault} & \textsc{Asteroids} & \textsc{MsPacman} & \textsc{Space Invaders}\\
		
		\rot{\qquad \quad Mean} &
		\includegraphics[width=40mm]{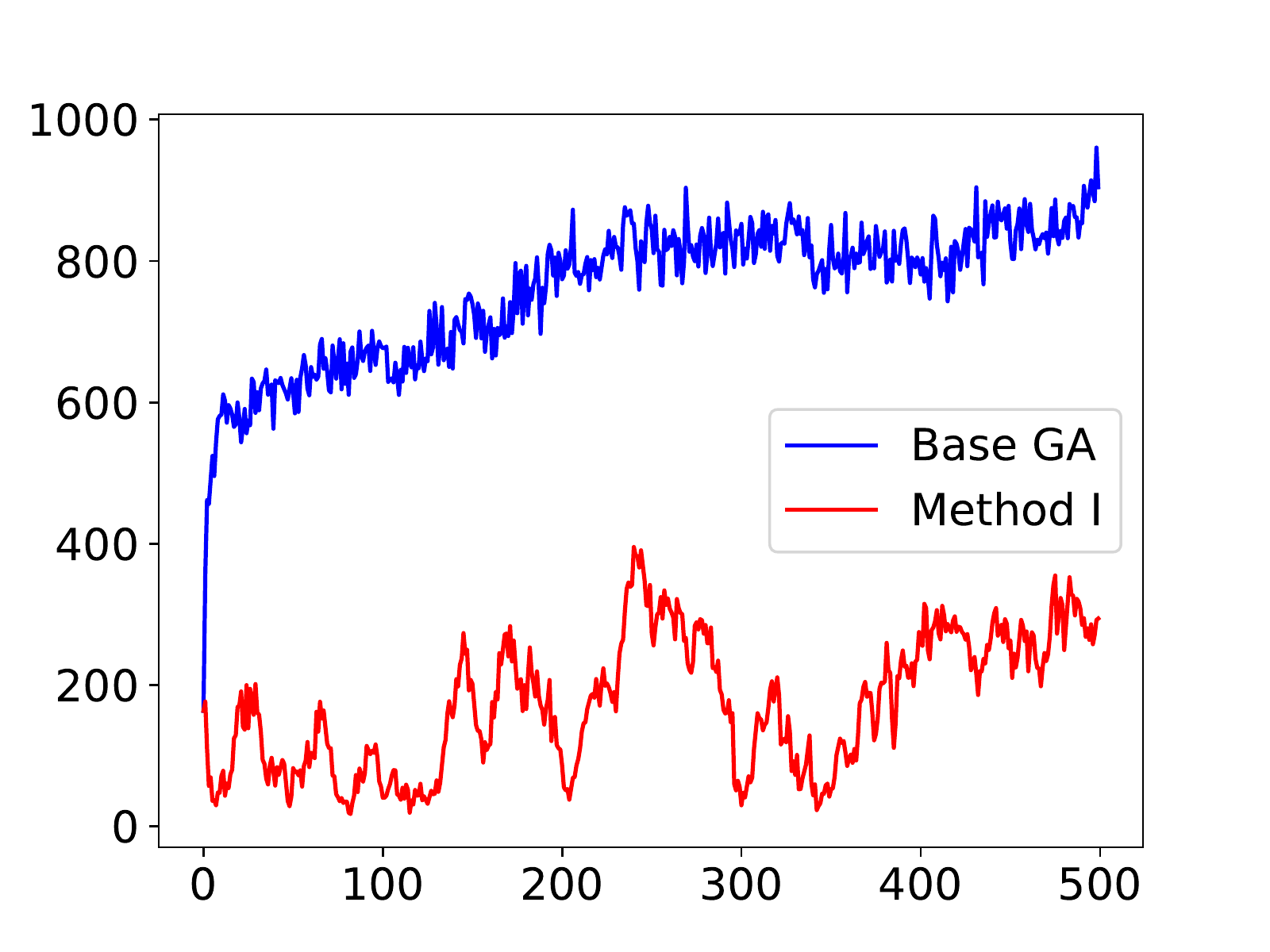} &   
		\includegraphics[width=40mm]{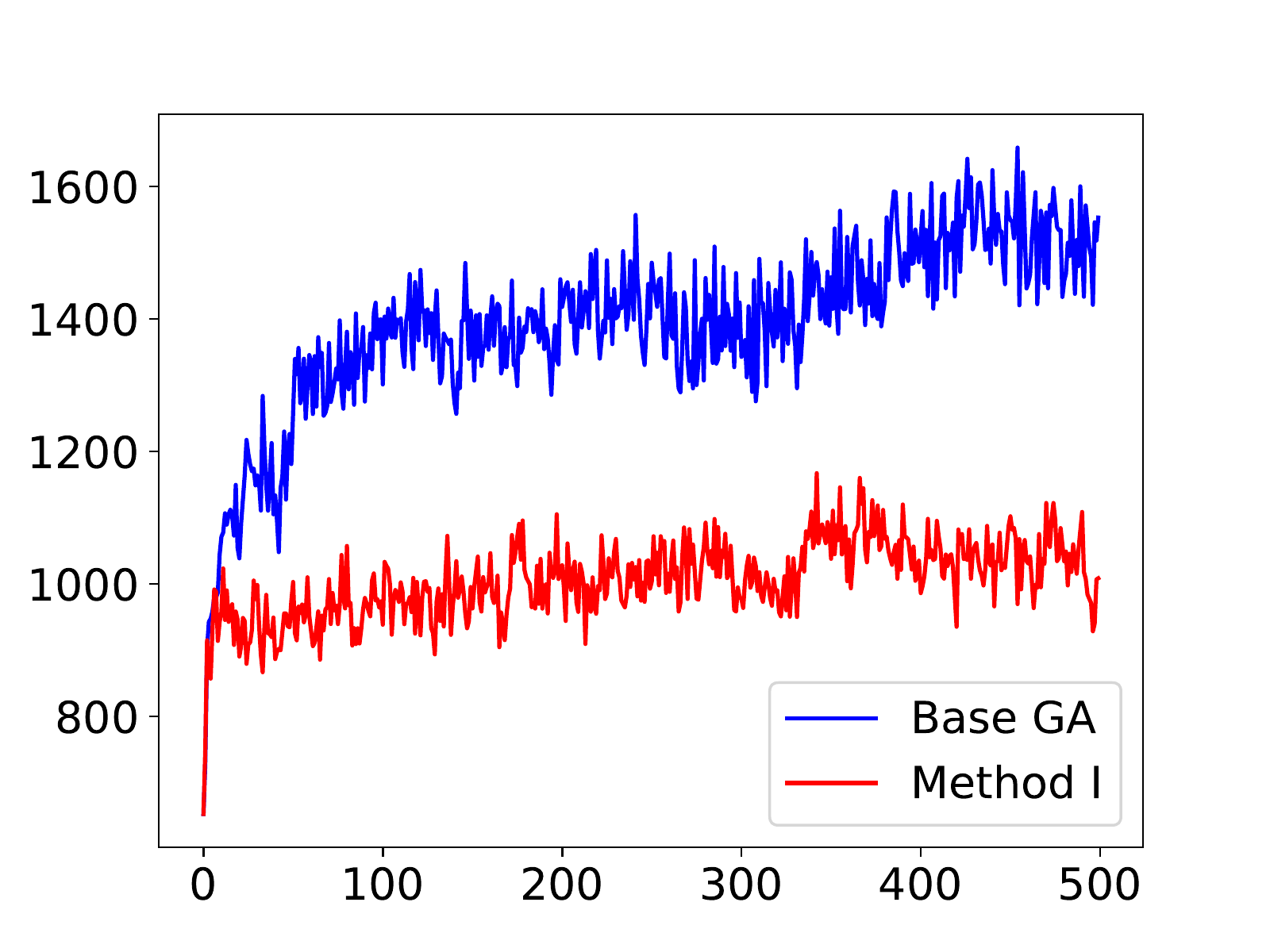} & 
		\includegraphics[width=40mm]{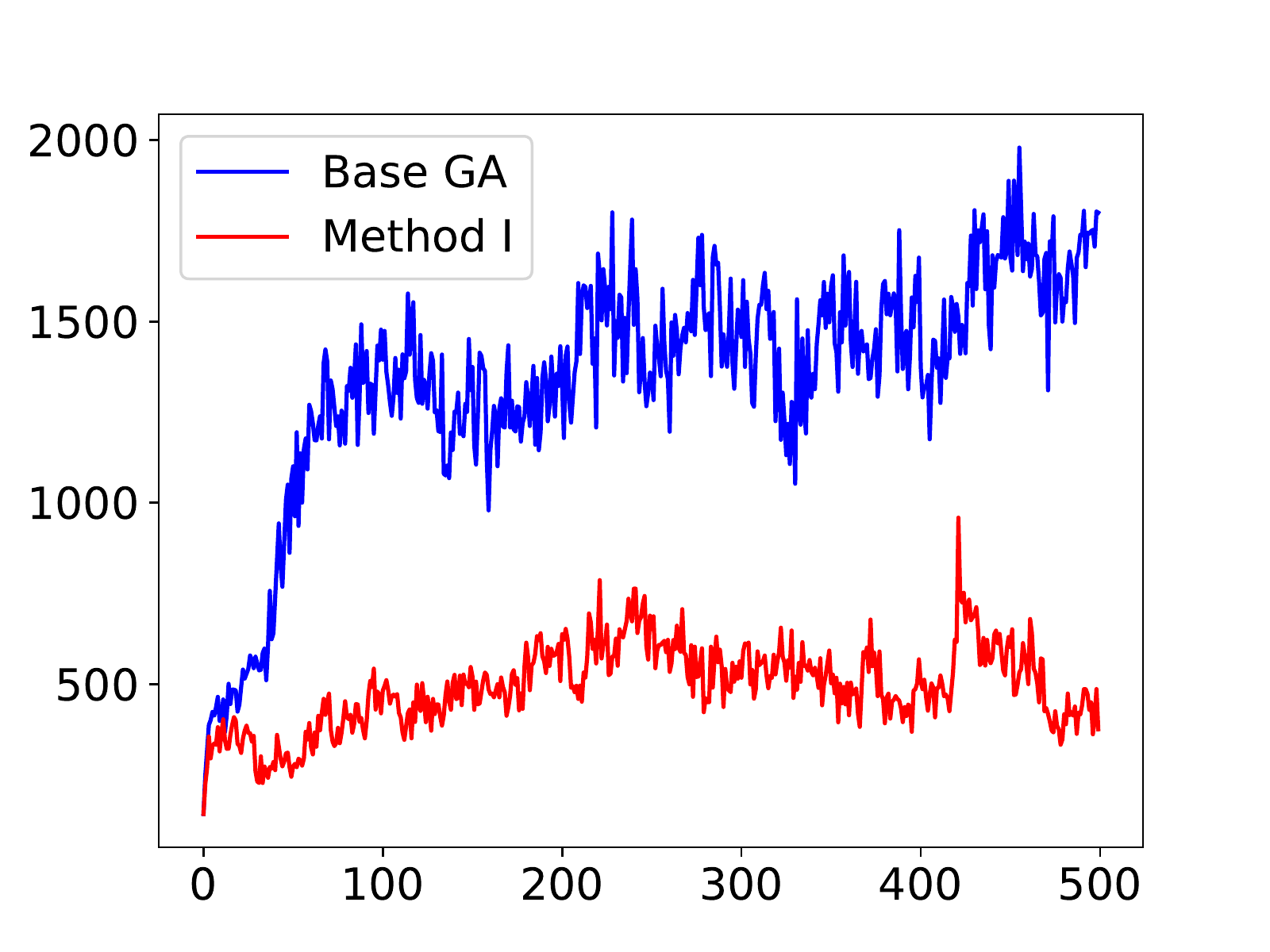} & 
		\includegraphics[width=40mm]{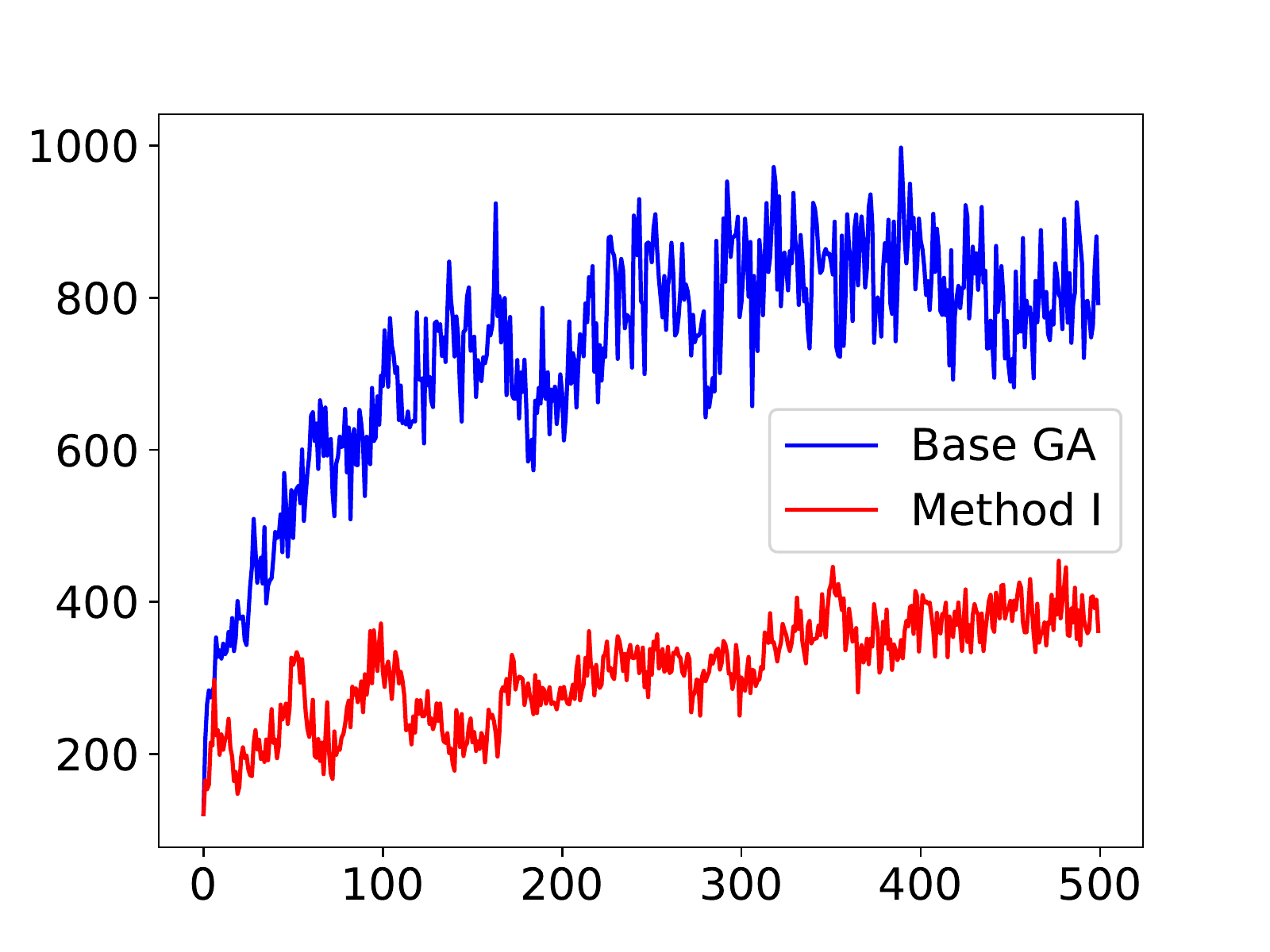} \\
		\rot{\qquad \quad High} &
		\includegraphics[width=40mm]{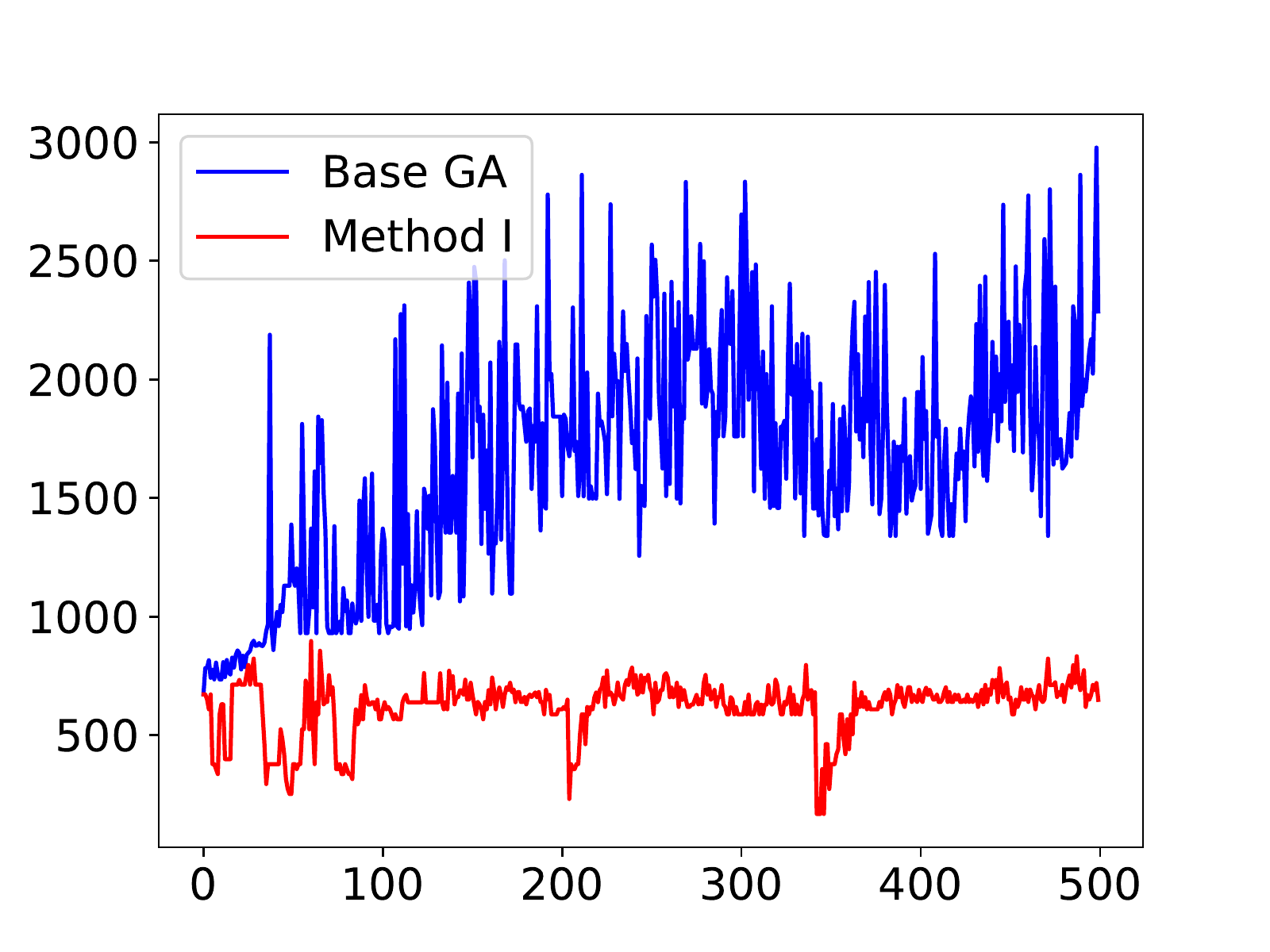}  &
		\includegraphics[width=40mm]{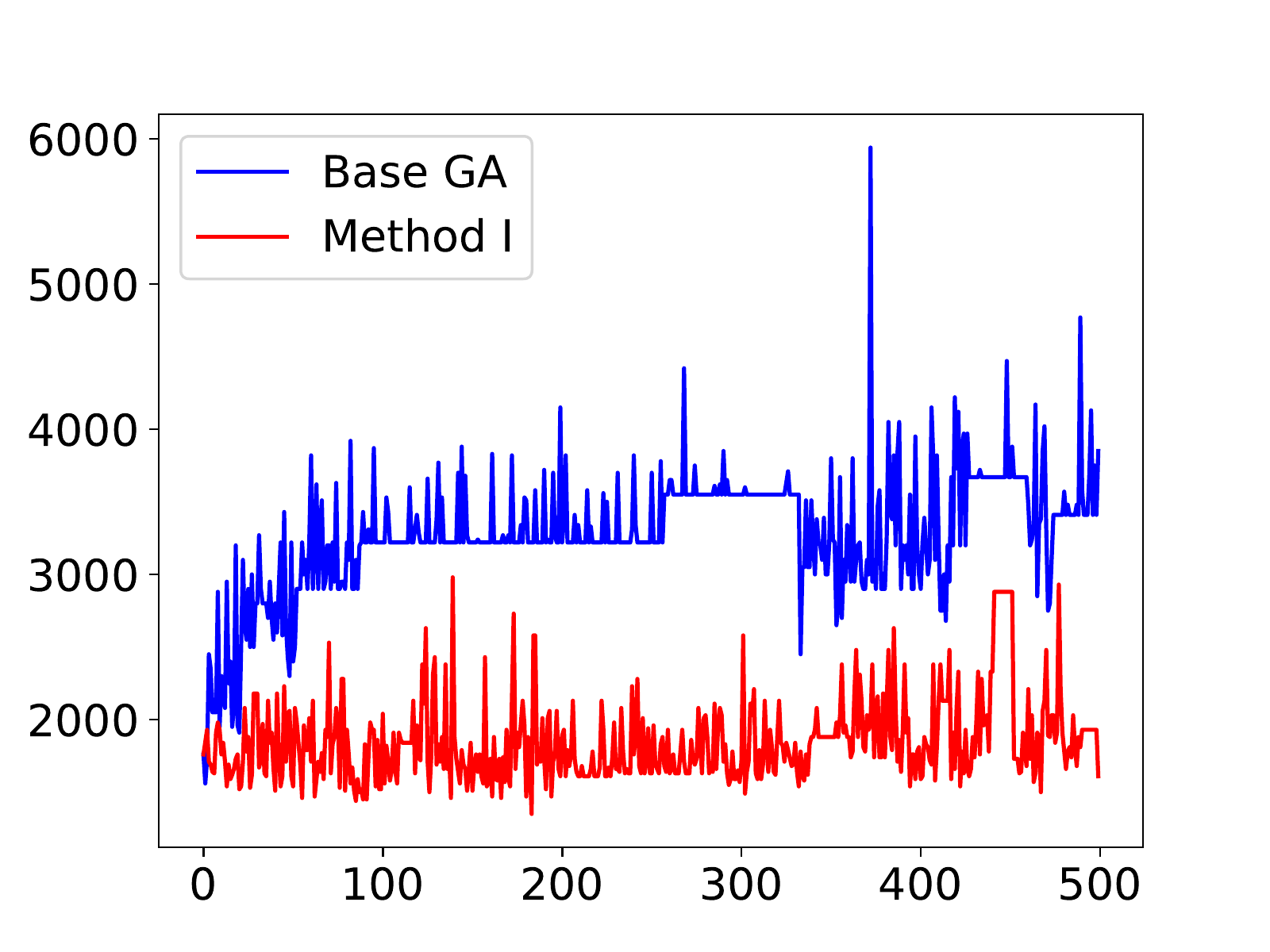}  &
		\includegraphics[width=40mm]{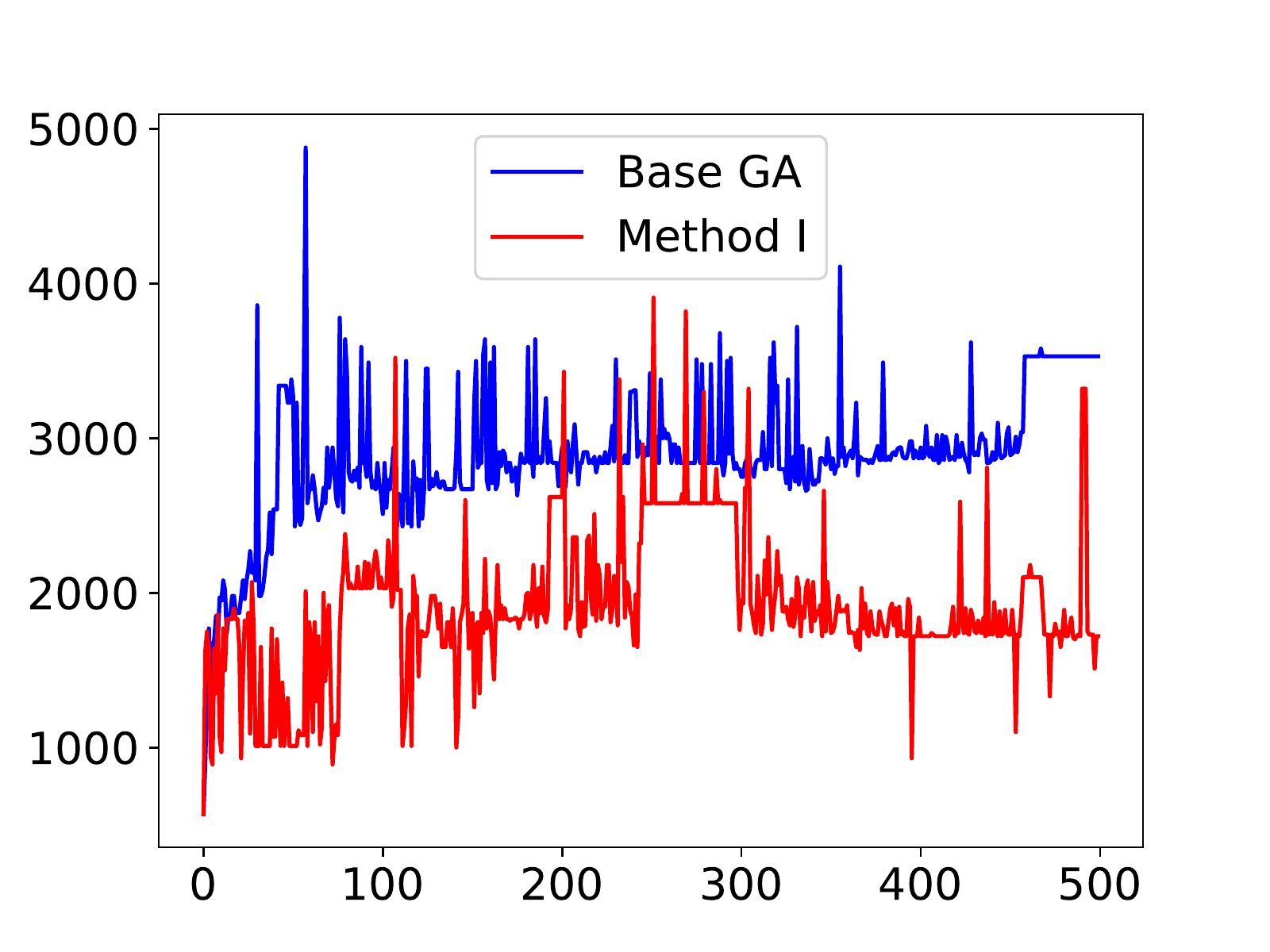}  &
		\includegraphics[width=40mm]{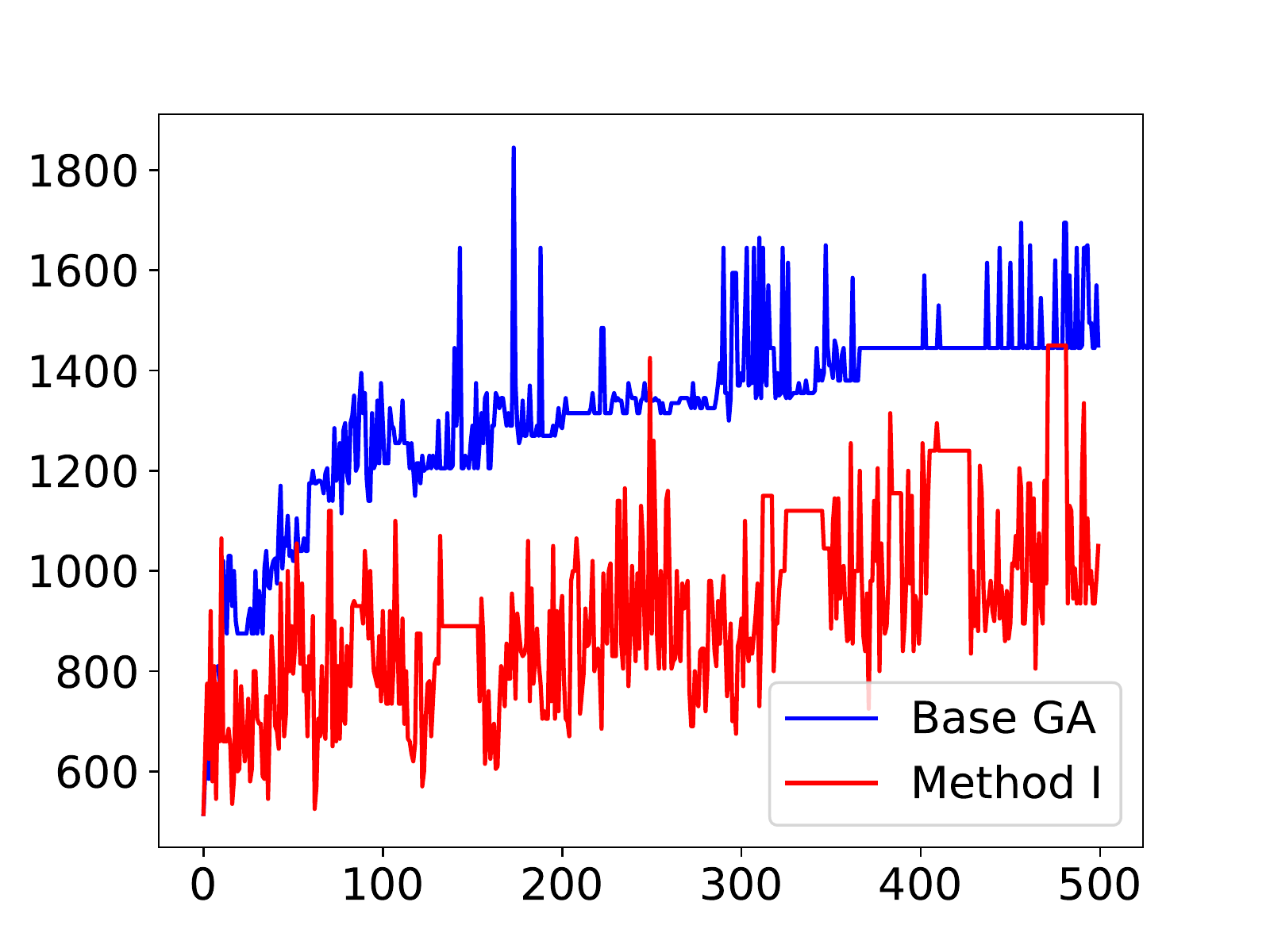}\\	
		\rot{\qquad Validation} &
		\includegraphics[width=40mm]{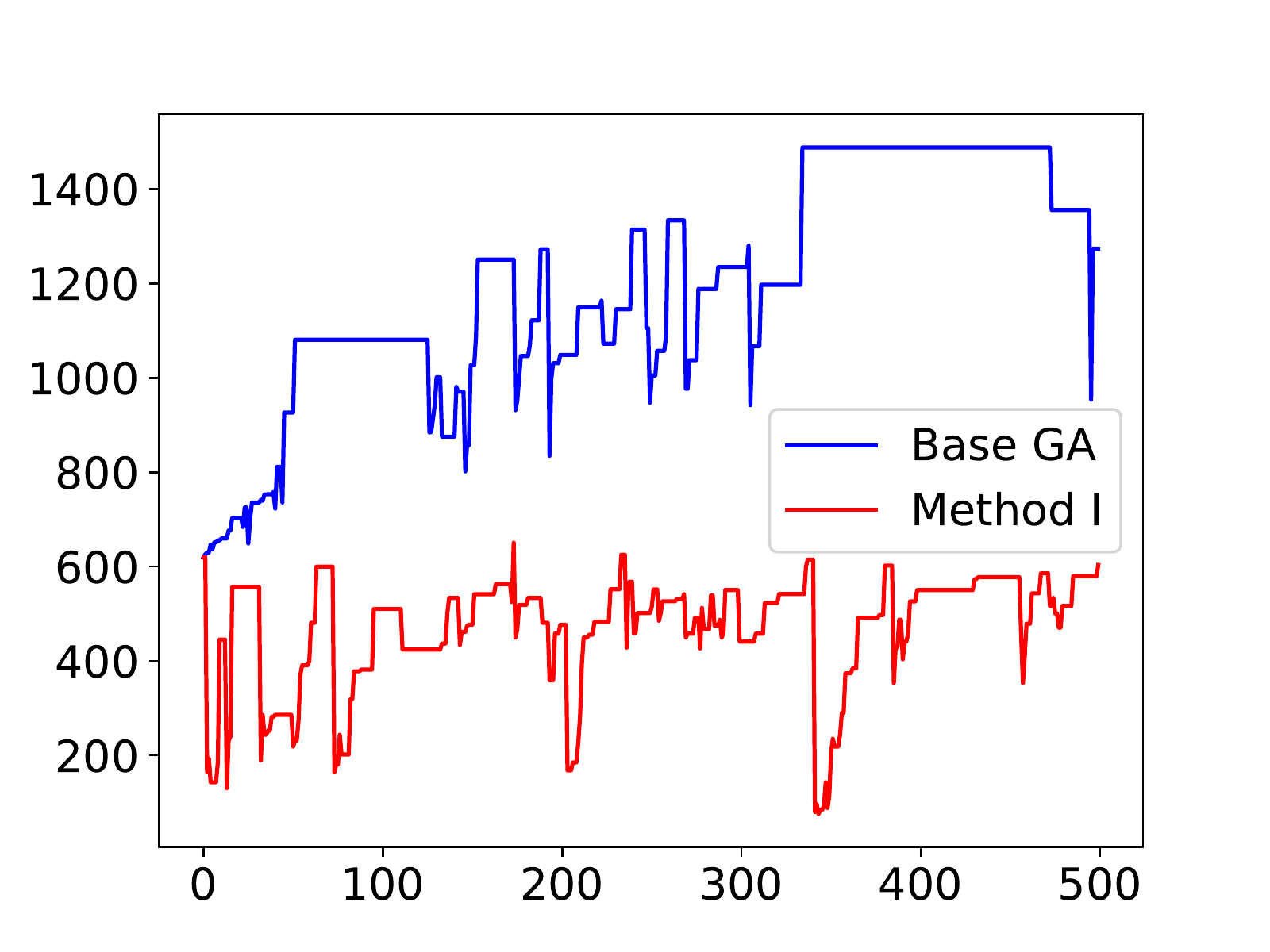}&
		\includegraphics[width=40mm]{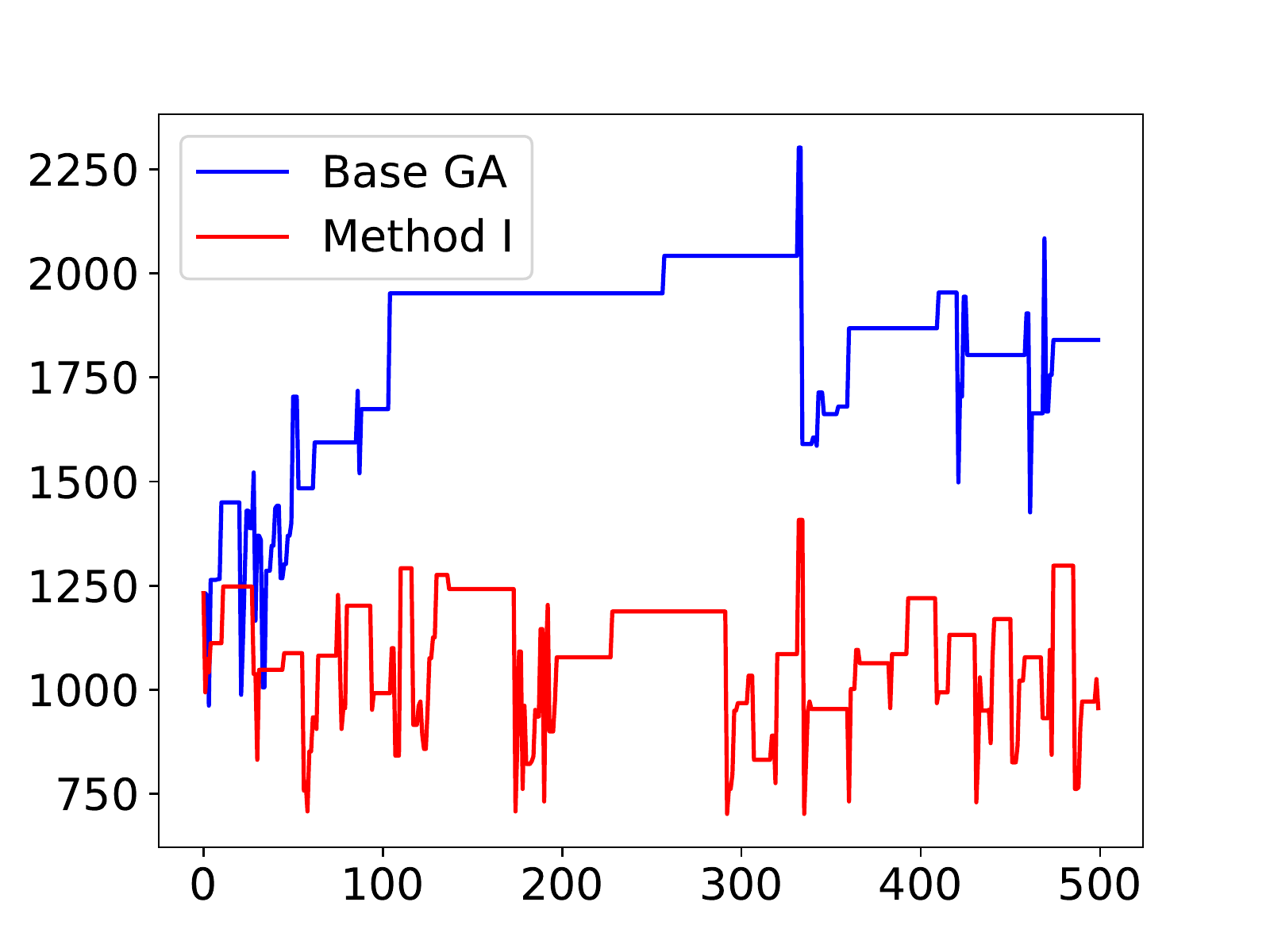}&
		\includegraphics[width=40mm]{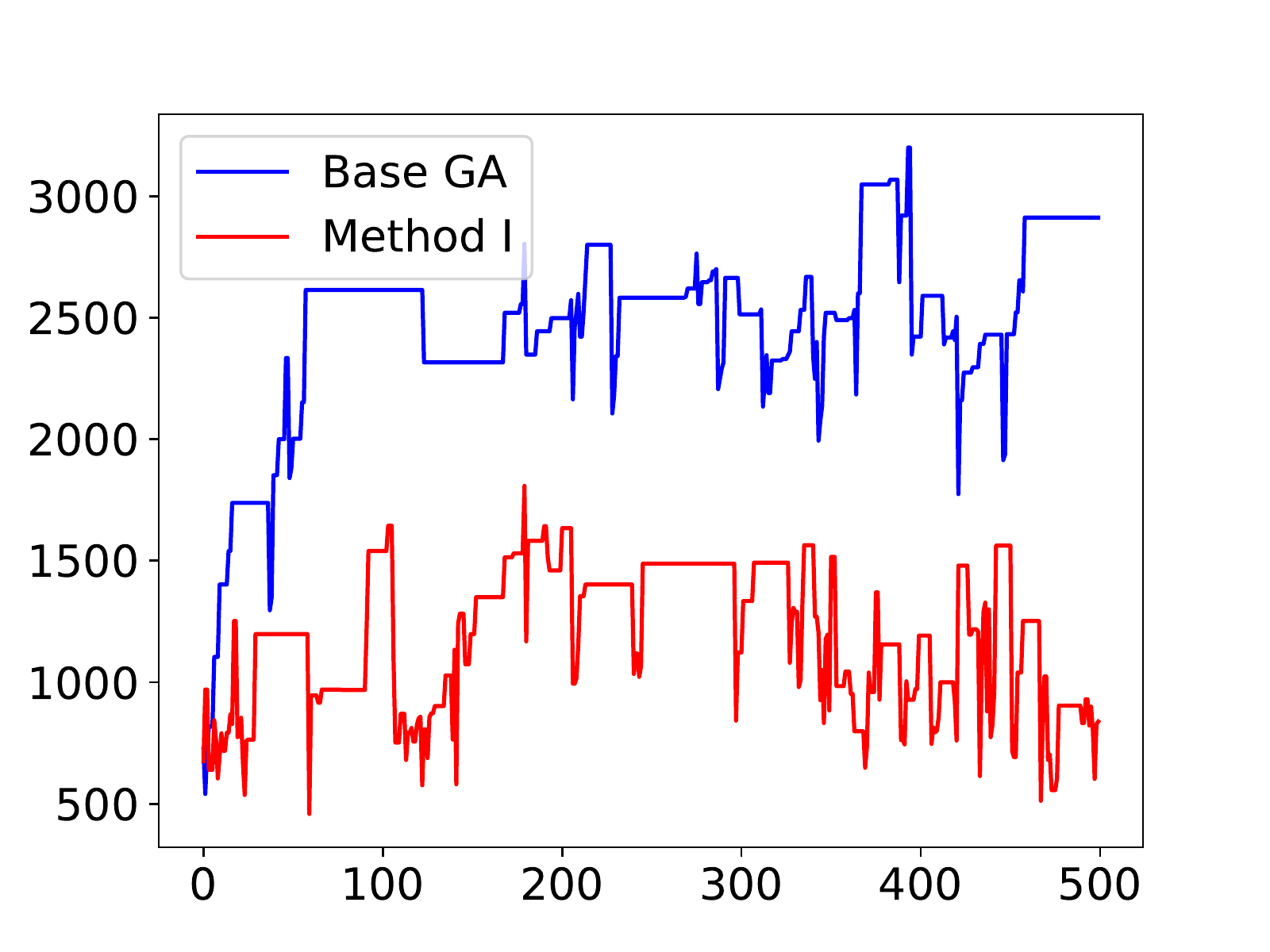}&
		\includegraphics[width=40mm]{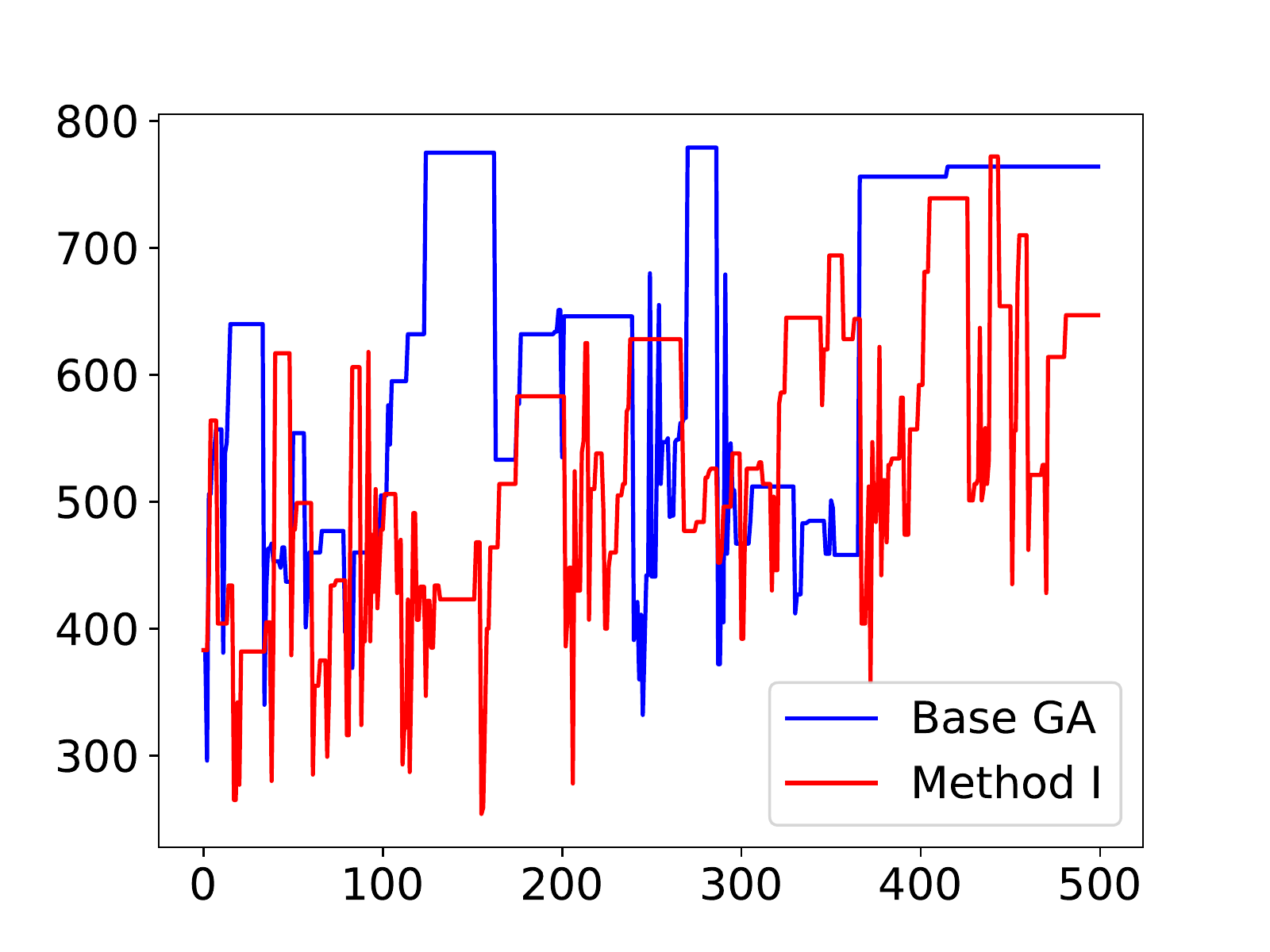}\\
	\end{tabular}
	\caption{Base GA and Method I learning progress.}
	\label{fig:method1-basega}
\end{figure*}

Method I training progress is visualized by Figure \ref{fig:method1-basega} and testing evaluation of Method I policies is summarized by Table \ref{table:method1results}. Overall, Method I does not produce policies that score better than either DQN or the Base GA. On the other hand, it is interesting to evaluate the behaviours of policies generated by (almost) completely ignoring the reward signal during training.

\begin{table}
	\begin{center}
		
		\begin{tabular}{lrrrr}
			\hline 
			& \multicolumn{2}{c}{ Mean} &   \multicolumn{2}{c}{ St.\ Dev.}     \\ 
			Game &  Base GA &  Method I  &  Base GA  &  Method I    \\
			\hline 
			\textsc{Assault} & \textbf{812} & 488 & 228 & 158  \\
			\textsc{Asteroids} & \textbf{1321} & 736 & 503 & 426   \\
			\textsc{MsPacman} & \textbf{2325} & 1437 & 351 & 527  \\
			\textsc{Space Invaders} & 500 & 474 & 303 & 195   \\
			\hline 
		\end{tabular} 
	\end{center}
	\caption{Comparison of Base GA and Method I testing results over 30 episodes not used in training or validation. Means and standard deviations are measured in game score units. Bolded means denote significantly better testing performance (p < $0.05$ in a two-tailed t-test). The Base GA  outperforms Method I in all but one game.}
	\label{table:method1results}
\end{table}

Results for Method I experiments suggest that novelty search indeed creates selection pressure for innovation. For example in \textsc{Space Invaders}, we observed more regular uses of obstacles by agents trained using Method I than the Base GA. And in \textsc{MsPacman}, we observed that agents trained using Method I tended to explore more paths than their Base GA-trained counterparts. In two out of four games (\textsc{Assault} and \textsc{Space Invaders}), agents trained by Method I had significantly longer lifespans than those trained by the Base GA (see Table \ref{table:method1lifespans}). This could either be due to the behavioural distance function's sensitivity to differences in lifespan, or to defensive innovations that increase agent lifespan. 

\begin{table}
	\begin{center}
		
		\begin{tabular}{lrrrr}
			\hline 
			& \multicolumn{2}{c}{ Mean} &   \multicolumn{2}{c}{ St.\ Dev.}     \\ 
			Game &  Base GA &  Method I  &  Base GA  &  Method I    \\
			\hline 
			\textsc{Assault} & 3538 & \textbf{5242} & 995 & 1998  \\
			\textsc{Asteroids} & 1263 & 1223 & 545 & 668   \\
			\textsc{MsPacman} & \textbf{1112} & 931 & 137 & 142  \\
			\textsc{Space Invaders} & 1264 & \textbf{1552} & 369 & 285   \\
			\hline 
		\end{tabular} 
	\end{center}
	\caption{Comparison of Base GA and Method I lifespans over 30 episodes not used in training or validation. Means and standard deviations are shown in numbers of frames over which agents survived. Bolded means denote significantly longer lifespans (p < $0.05$ in a two-tailed t-test). Method I produced agents with significantly longer mean lifespans in testing in \textsc{Assault} and \textsc{Space Invaders}.}
	\label{table:method1lifespans}
\end{table}

\begin{figure}

			\includegraphics[width=50mm]{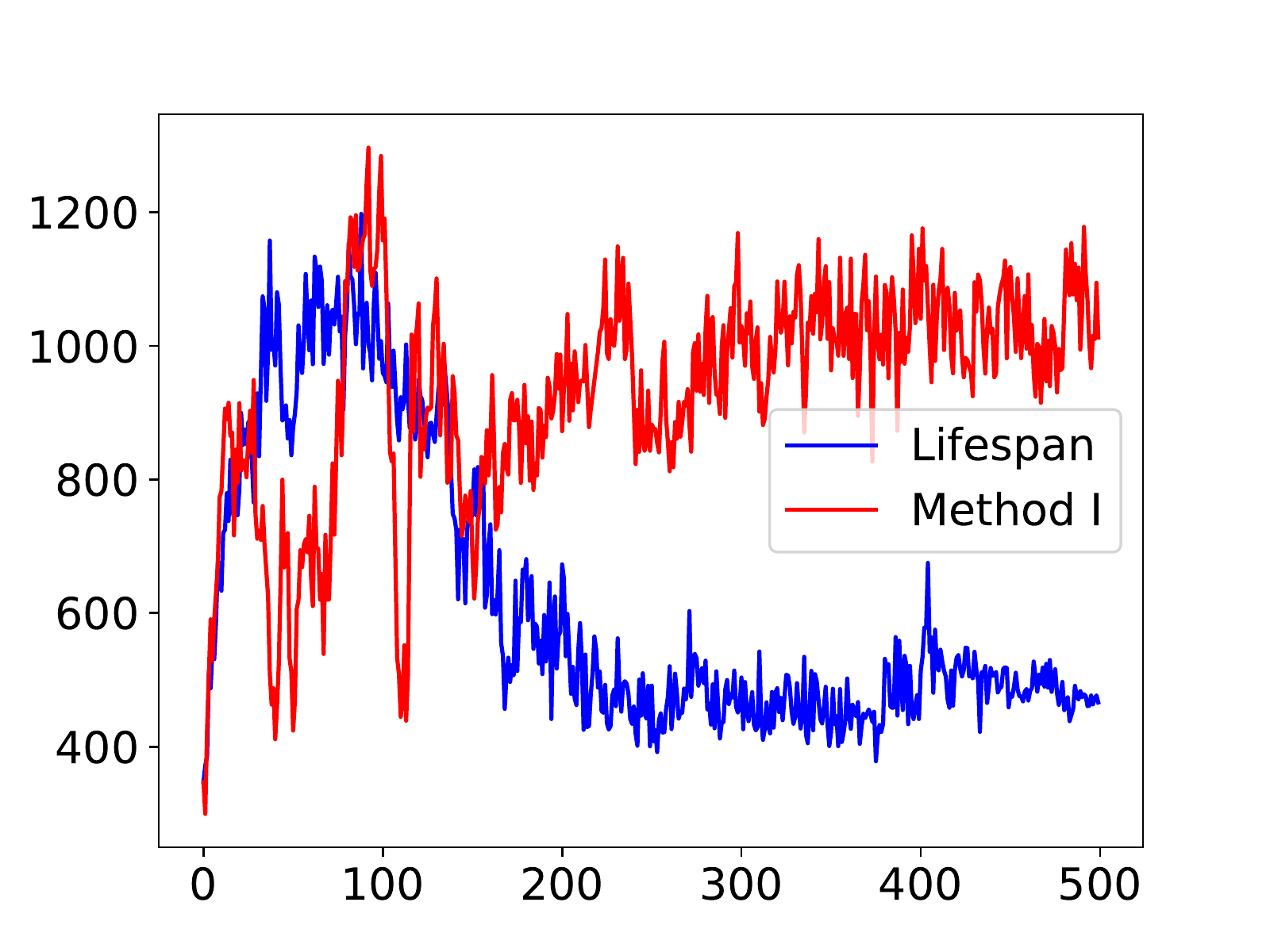}%
			\caption{Population mean game score over generations during training on \textsc{MsPacman}. Mean scores diverge after generation 160. Levenshtein distance (Method I) and lifespan are thus not equivalent behavioural distance functions.}%
			\label{fig:lifespan}
\end{figure}

To determine whether Levenshtein distance is effectively different than lifespan as a behavioural distance function, we conducted an additional small-scale experiment using \textsc{MsPacman} (Method I-L). We observed that lifespan, measured by counting the number of frames an agent survives in its environment, is not an equivalent behavioural distance function to Levenshtein distance. See Table \ref{table:lifespan} for hyperparameters and  Figure \ref{fig:lifespan} for results.

\begin{table}
			\begin{tabular}{lr}
				\hline
				Hyperparameter & Method I-L \\ 
				\hline 
				Population Size (N) & 100 + 1 \\ 
				
				Generations & 500  \\ 
				
				Truncation Size (T) & 10  \\ 
				
				Mutation Power ($\sigma$) & 0.004  \\ 
				
				Archive Probability & 0.02  \\ 
				
				Max Frames Per Episode (F) & 2,500  \\ 
				
				Training Episodes & 2  \\ 
				
				\hline 
			\end{tabular} 
			\caption{Hyperparameters for experiment on Method I-L. Validation episodes were not used; elites were determined using highest game score in training over 2 episodes. }%
			\label{table:lifespan}
		
\end{table}

A problem with this approach is that by continually selecting for innovation, there may be insufficient evolutionary time for innovations to be optimized. Method II attempts to remedy this by integrating secondary selection pressure for novelty into an otherwise standard search for reward-optimizing policies. 

\subsection{Method II}\label{sec:method2}

Method II was designed to help the GA avoid \textit{stagnation} or premature convergence to locally optimal solutions. This method adds two components to the Base GA: 1) stagnation detection, and 2) population resampling. Stagnation is detected by examining the trend of validation scores. In the Base GA, validation episodes are used solely to identify the elite individual of a population. In Method II, learning progress is declared to be stagnant when validation scores are non-increasing over 10 episodes. This is reflected by the hyperparamter \textit{Improvement Generations} (\emph{IG}) in Table \ref{table:method2params}. Population resampling is achieved by sampling $2*T$ individuals from the archive to be the next generation's parents. For Method II, novelty scores are used to select  archived individuals whose policies were most different from the current population, according to the behavioural distance metric. 

As a baseline, we tested whether novelty-based population resampling is better than sampling random individuals from the archive for learning \textsc{MsPacman} policies. Using the same evaluation criteria and hyperparameters as for Method II (see Table \ref{table:method2params}) we found that, for \textsc{MsPacman}, novelty-based population resampling is significantly better than random archive sampling. This result is summarized by Table \ref{table:randomresults} and motivated further evaluation of the method applied to other games. 

\begin{figure*}
	\begin{tabular}{ccccc}
		\multicolumn{5}{c}{Method II Learning Progress}\\
		\hline
		& \textsc{Assault} & \textsc{Asteroids} & \textsc{MsPacman} & \textsc{Space Invaders}\\
		
		\rot{\qquad \quad Mean} &
		\includegraphics[width=40mm]{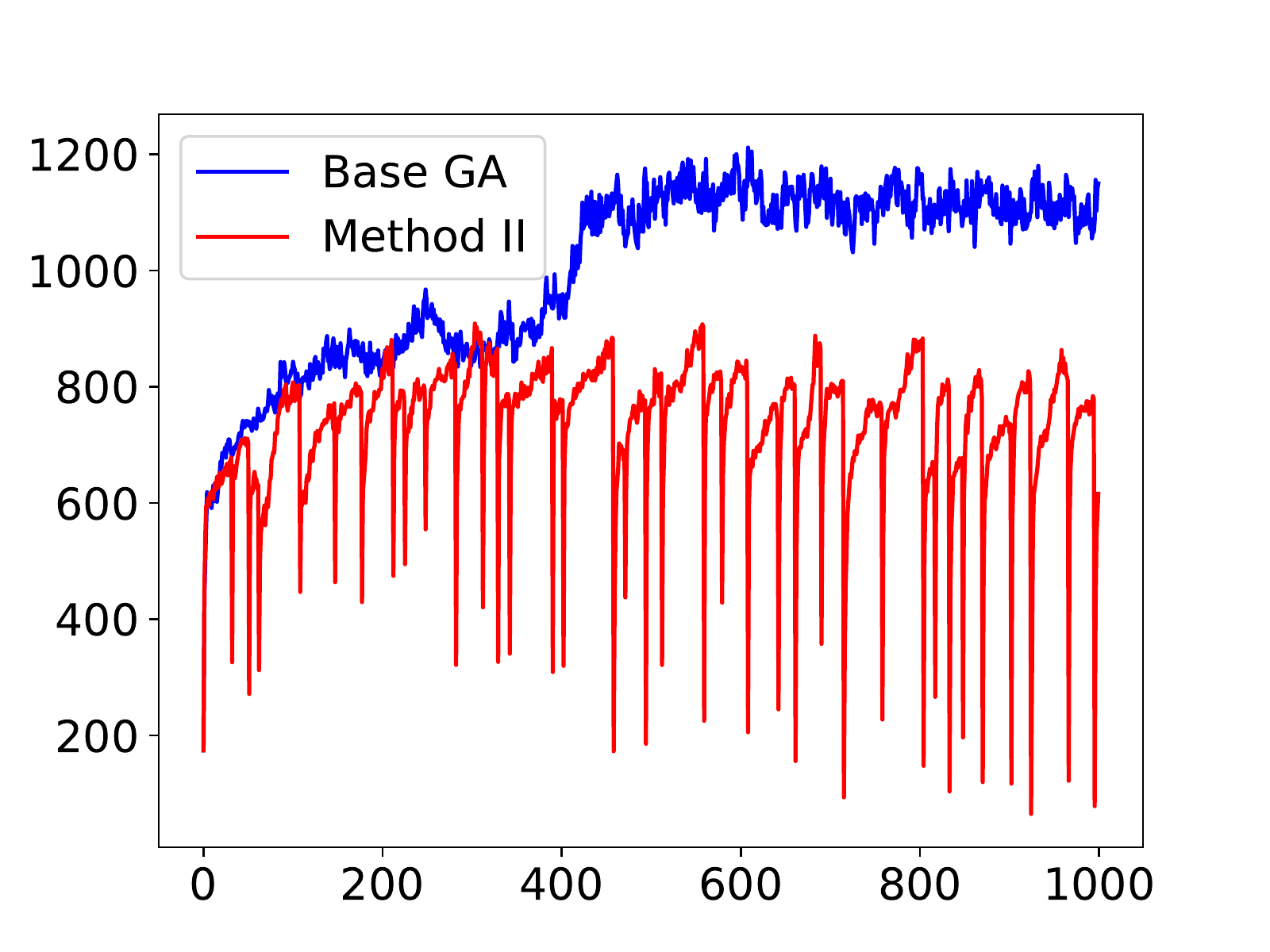} &   
		\includegraphics[width=40mm]{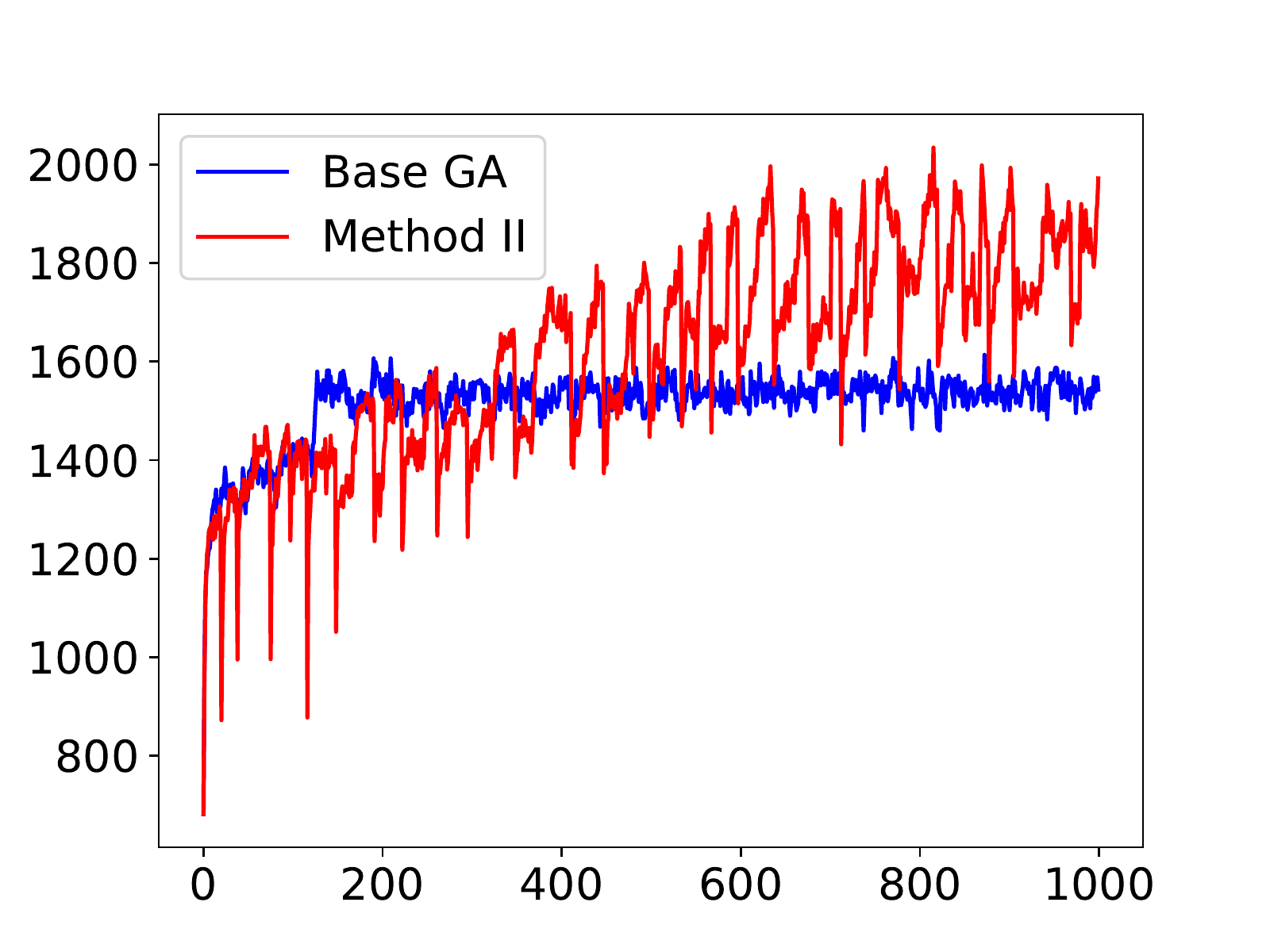} & 
		\includegraphics[width=40mm]{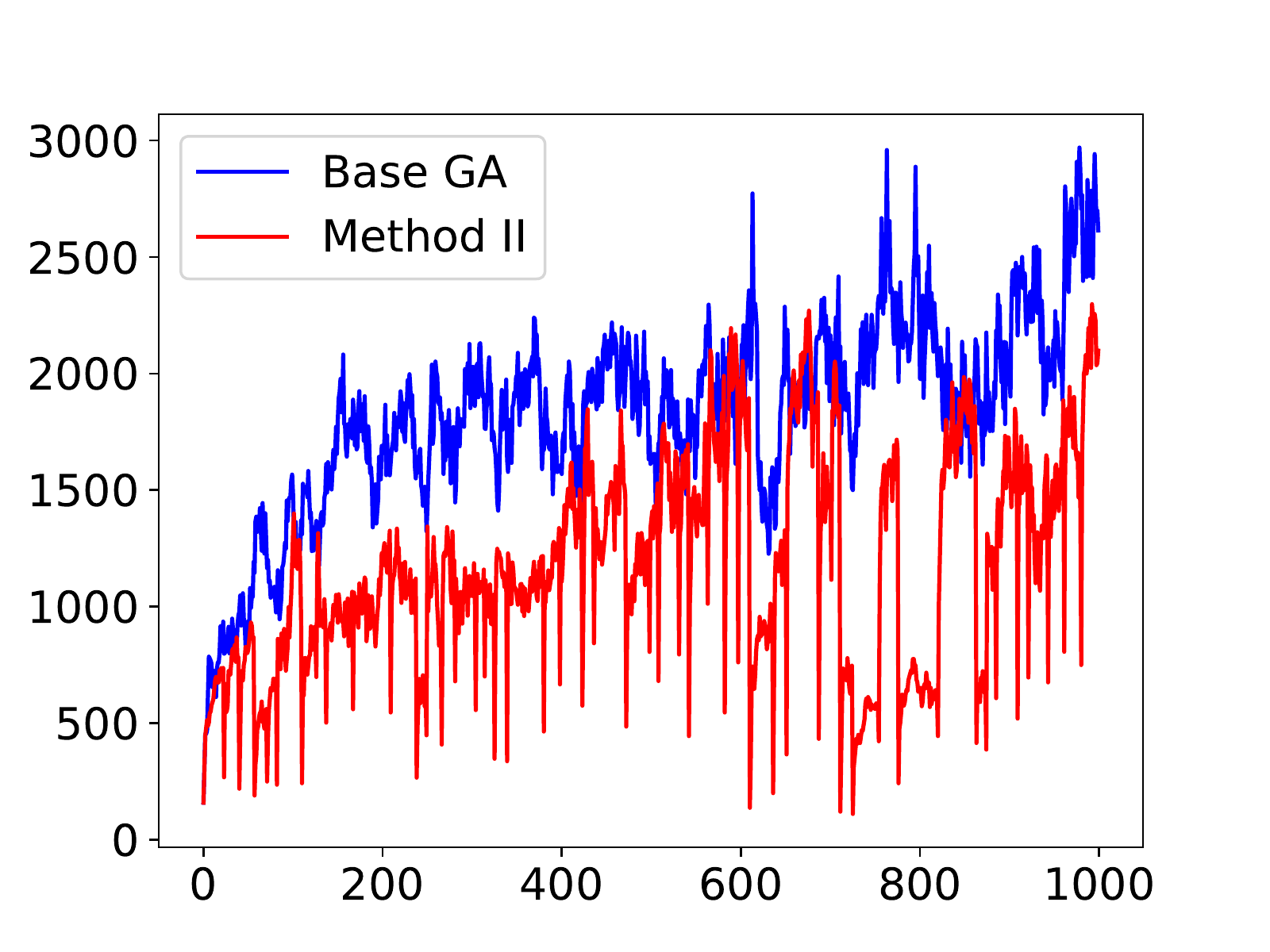} & 
		\includegraphics[width=40mm]{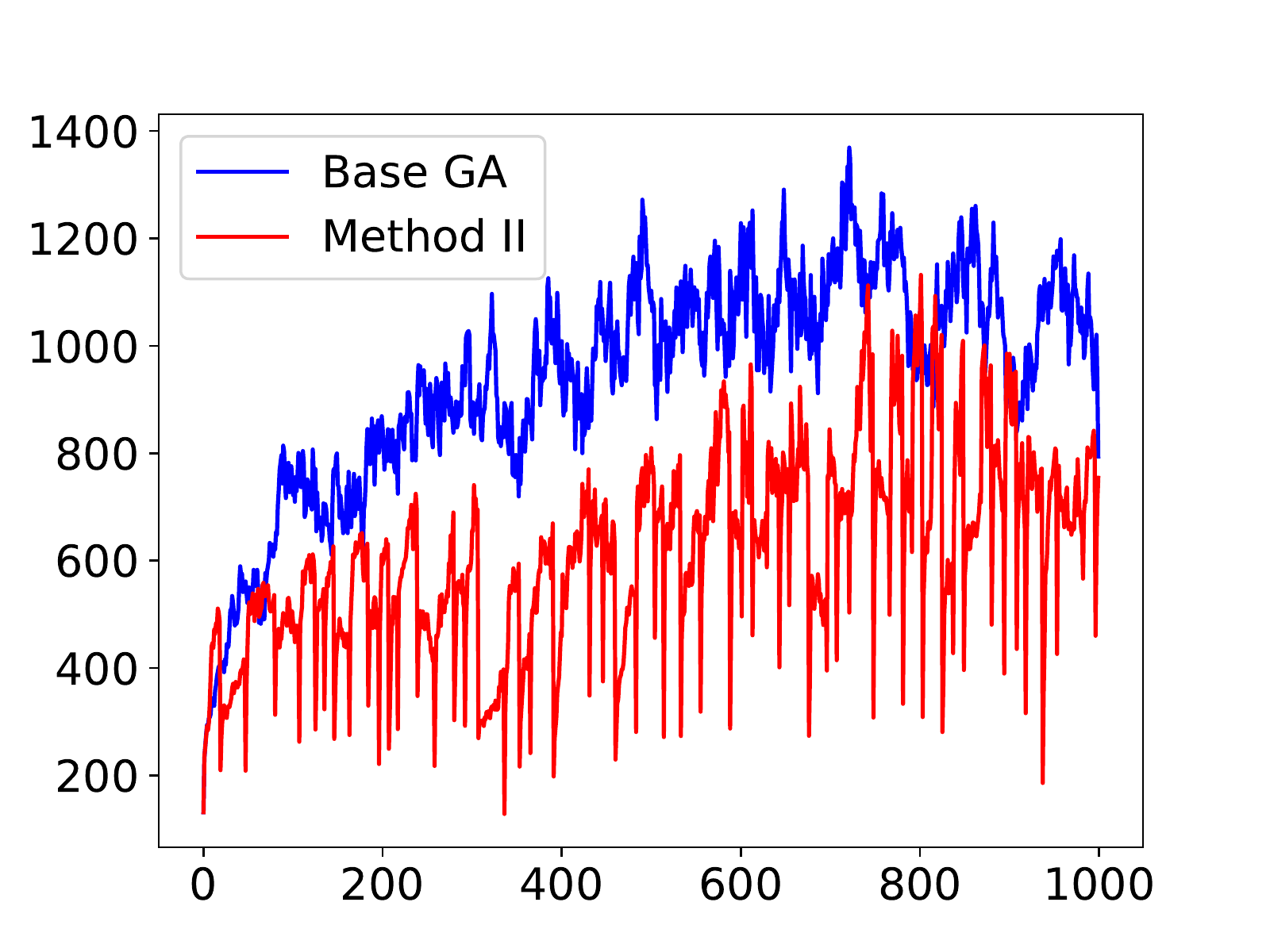} \\
		\rot{\qquad \quad High} &
		\includegraphics[width=40mm]{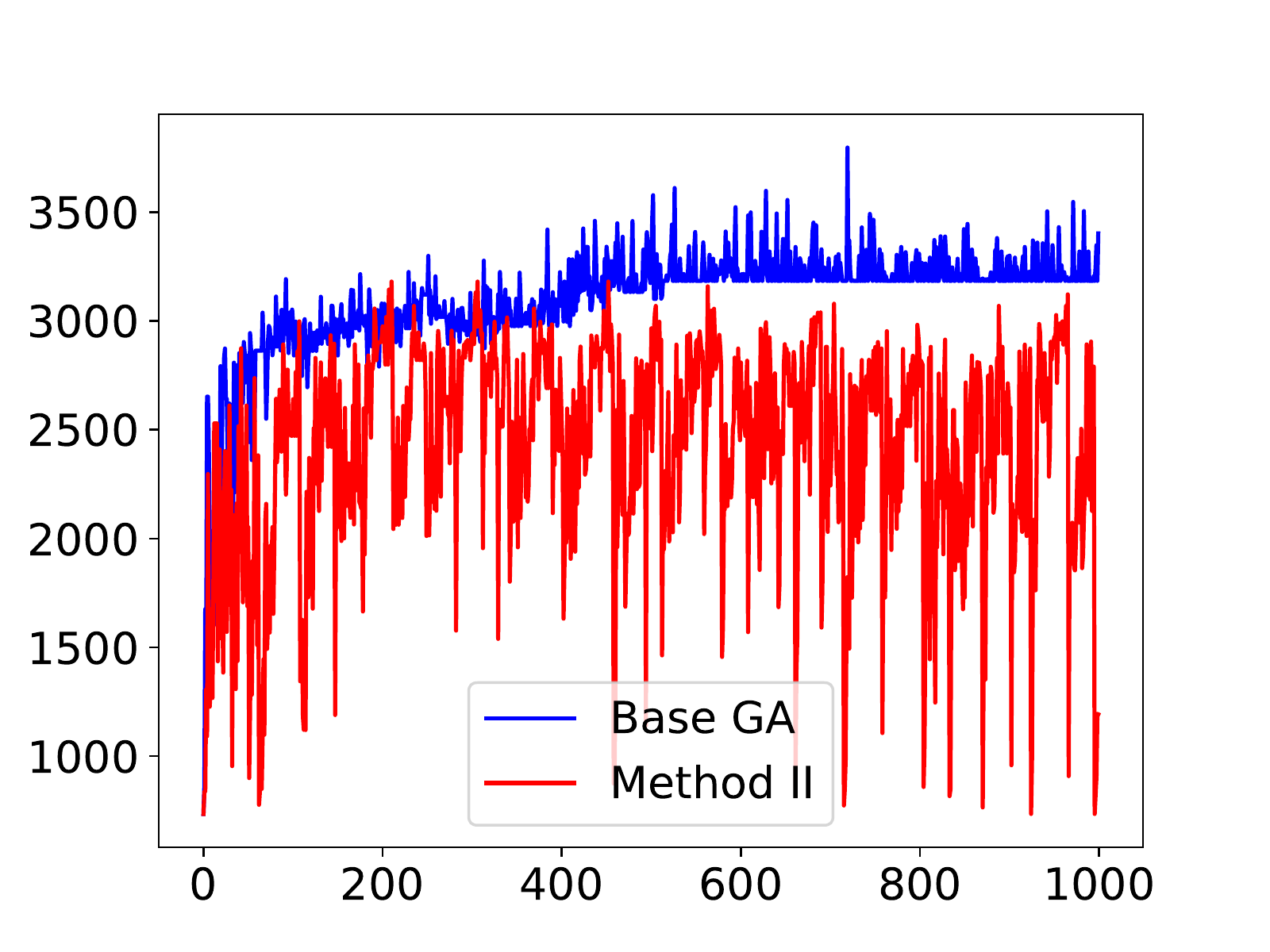}  &
		\includegraphics[width=40mm]{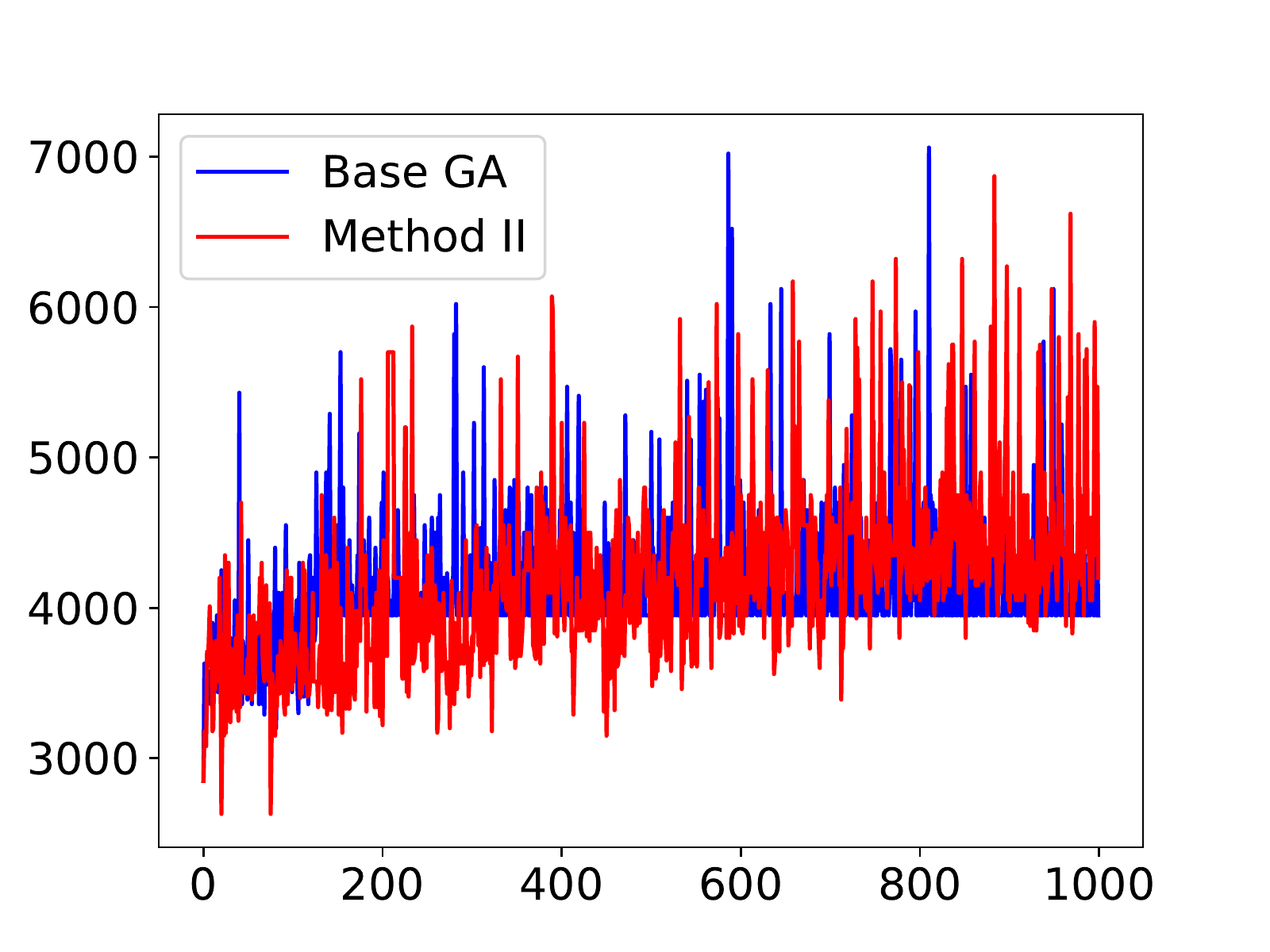}  &
		\includegraphics[width=40mm]{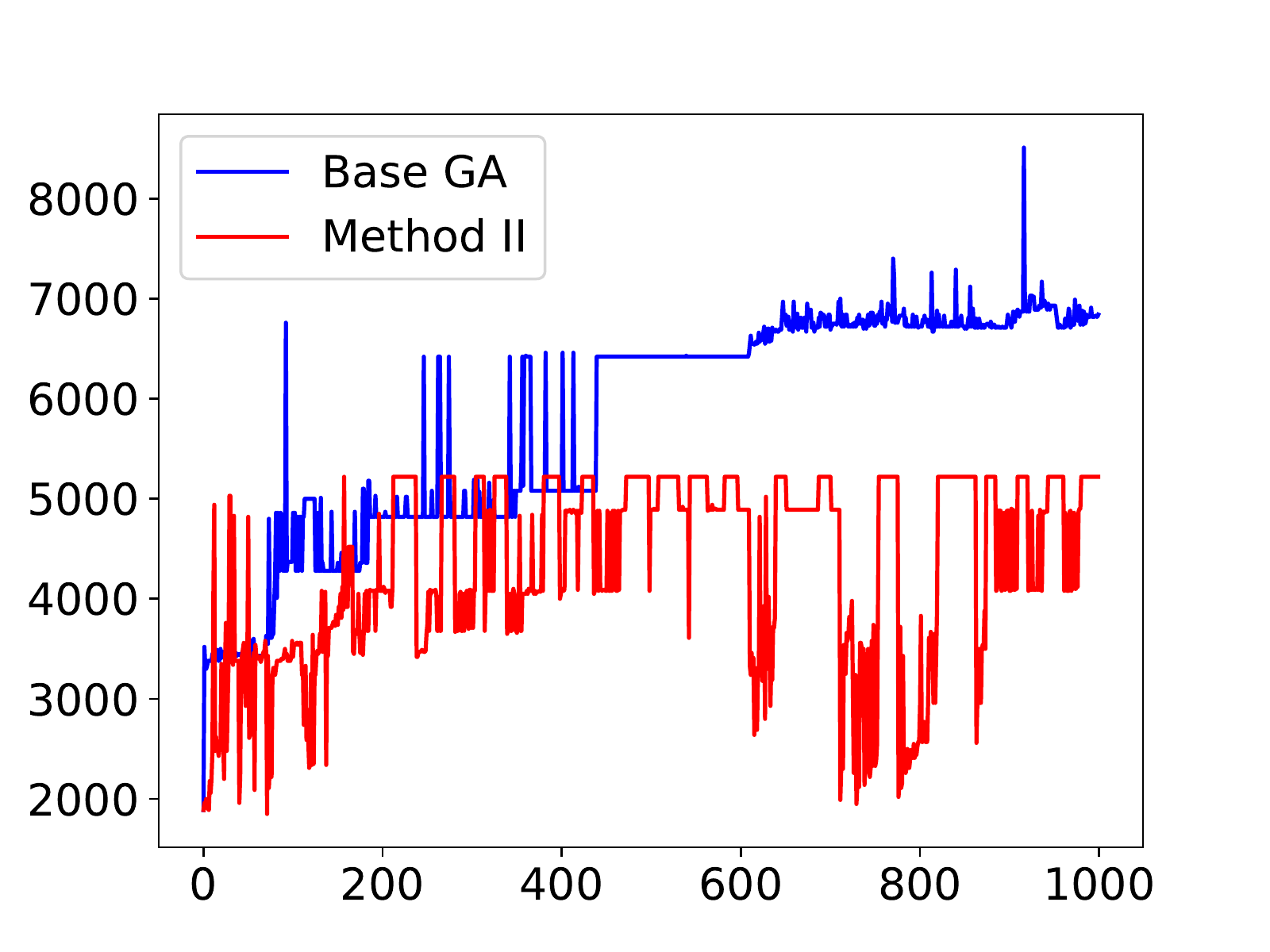}  &
		 \includegraphics[width=40mm]{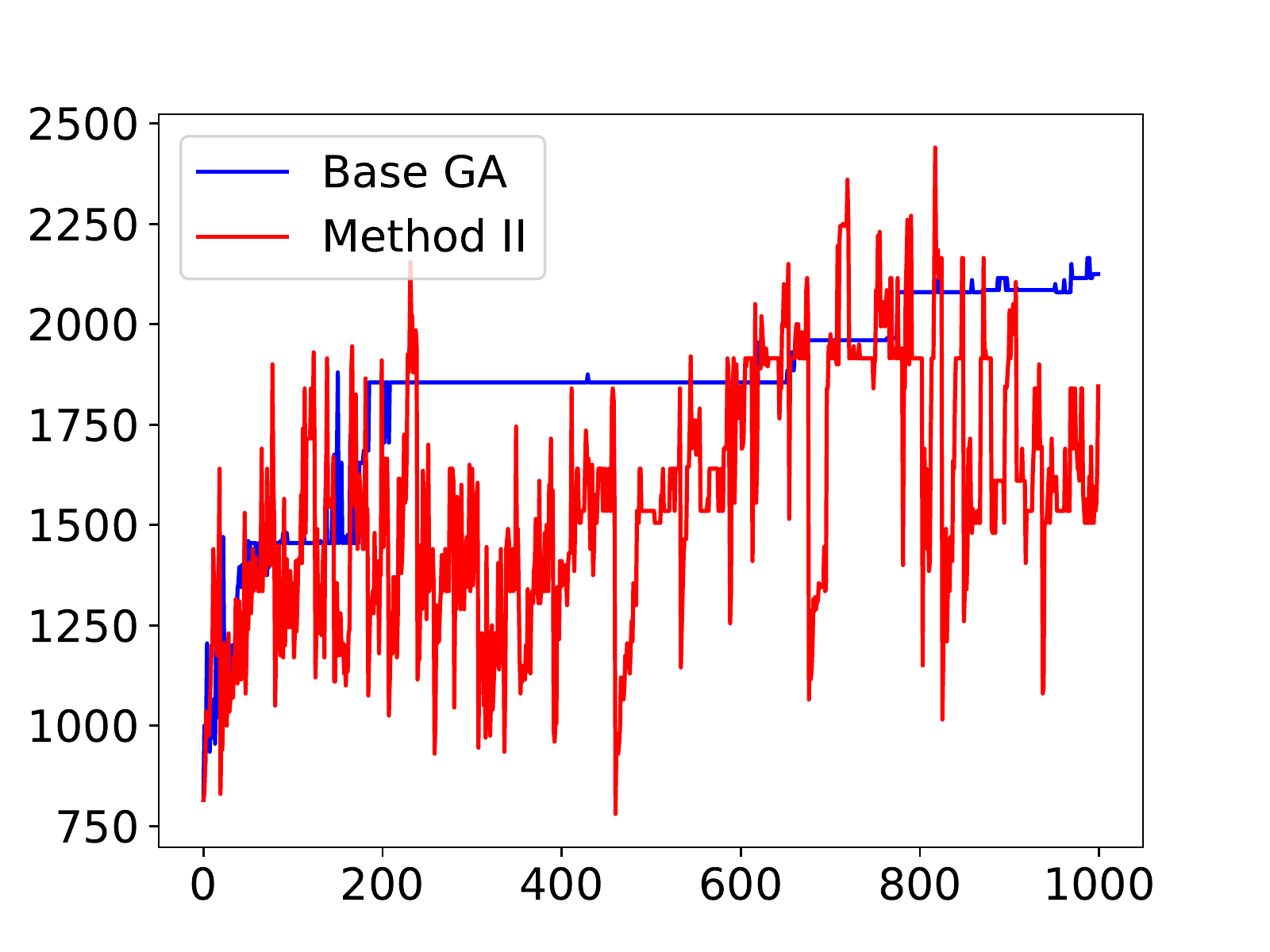}\\	
		\rot{\qquad Validation} &
		\includegraphics[width=40mm]{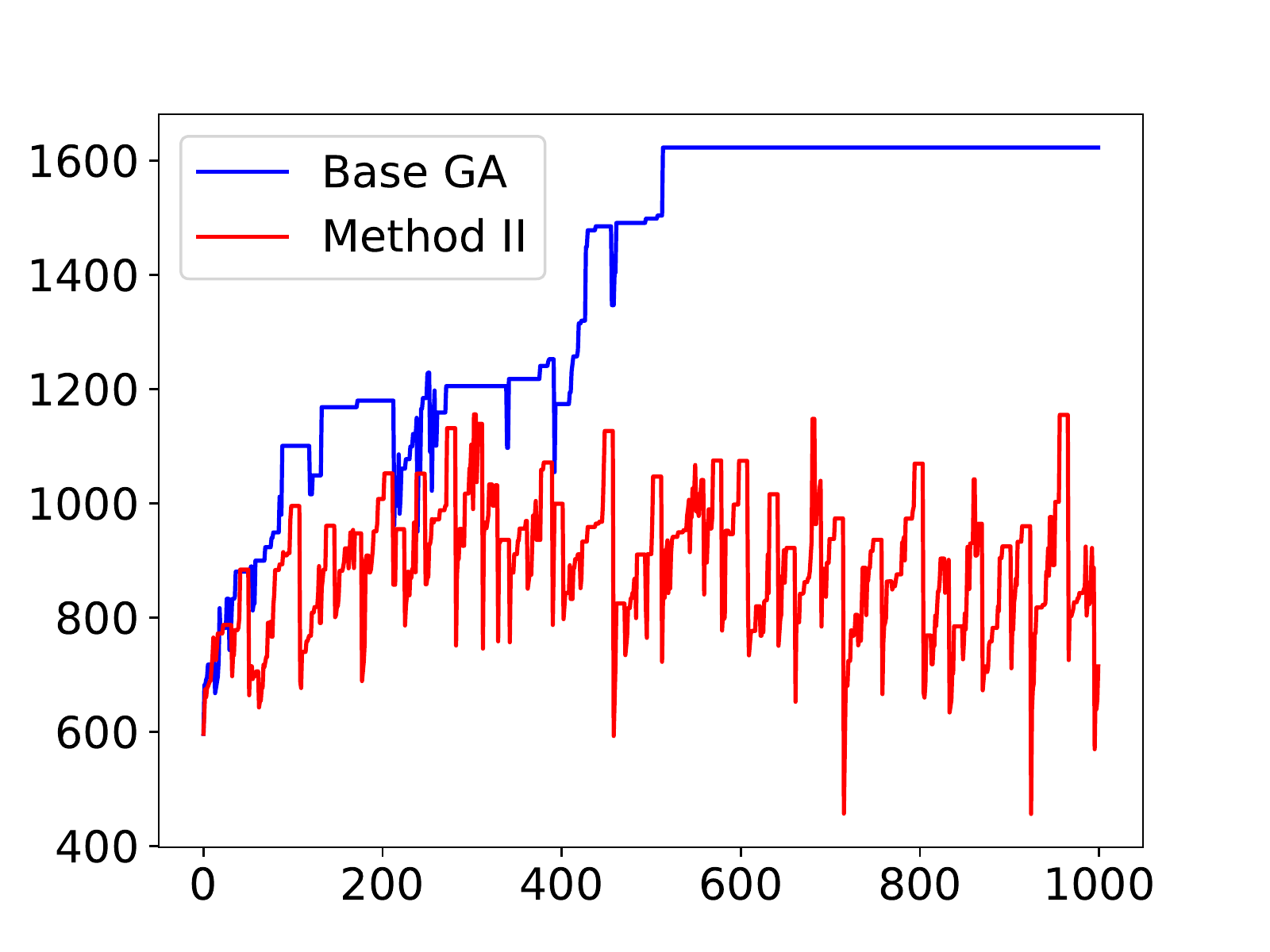}&
		\includegraphics[width=40mm]{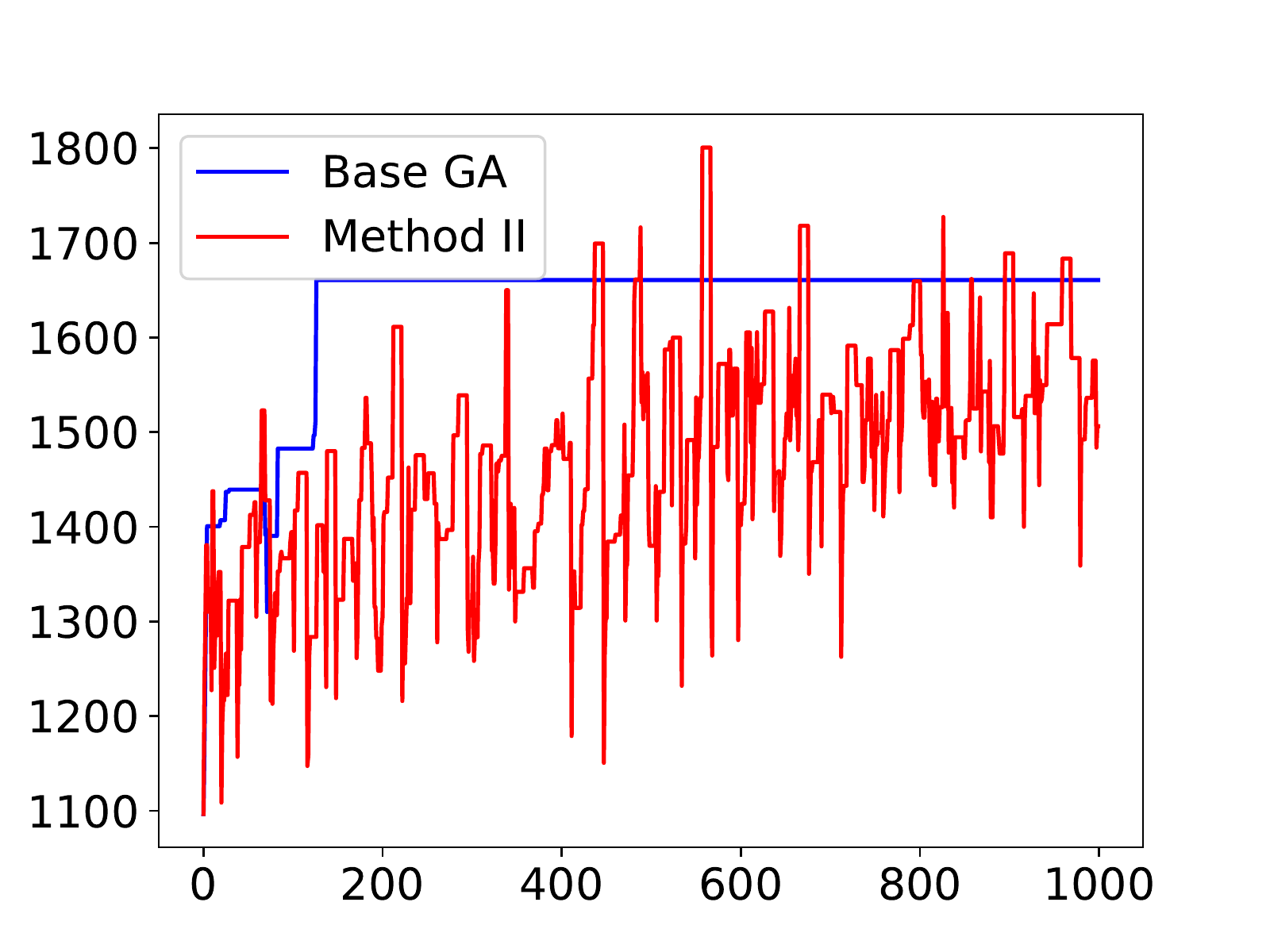}&
		\includegraphics[width=40mm]{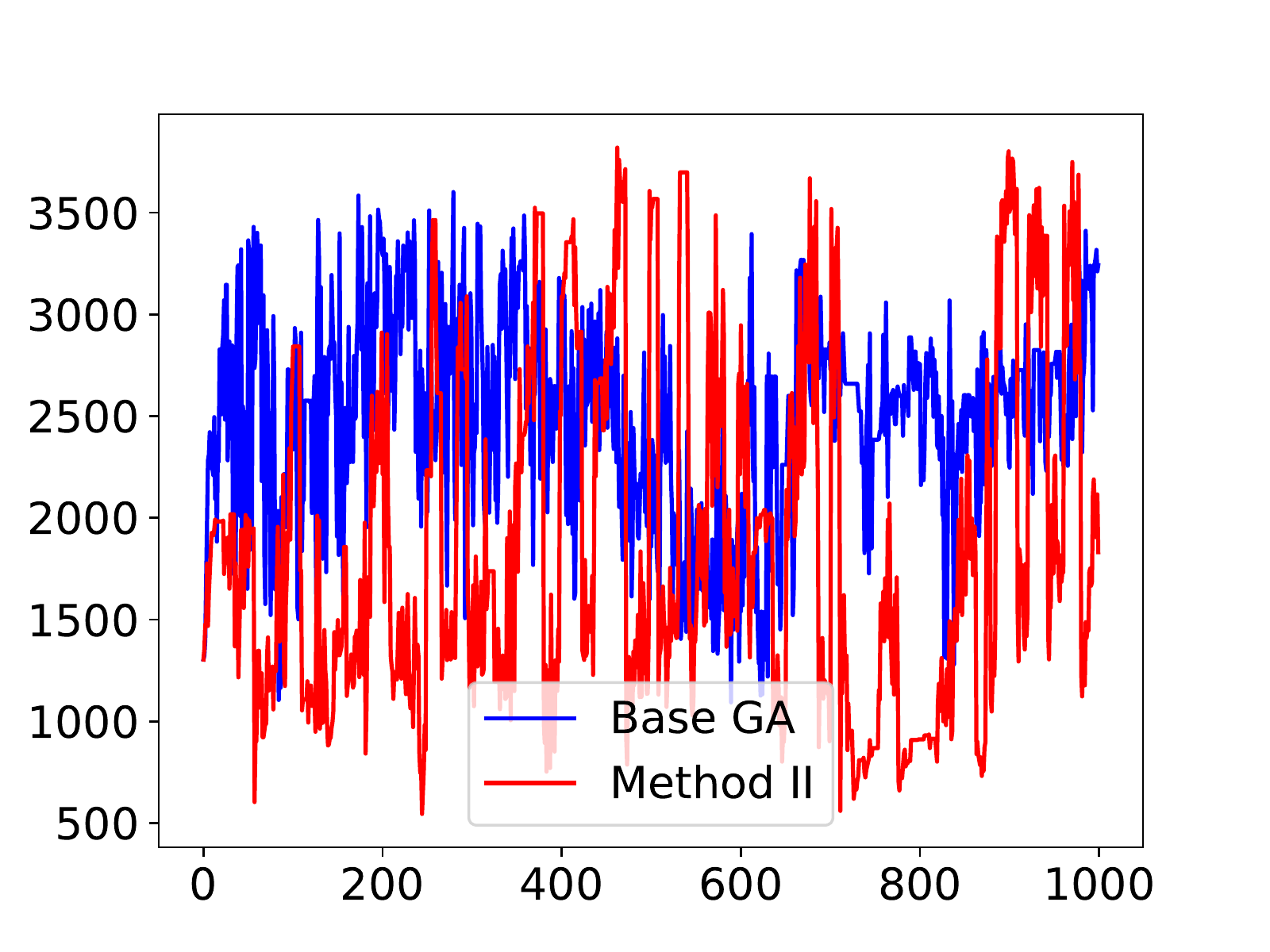}&
		\includegraphics[width=40mm]{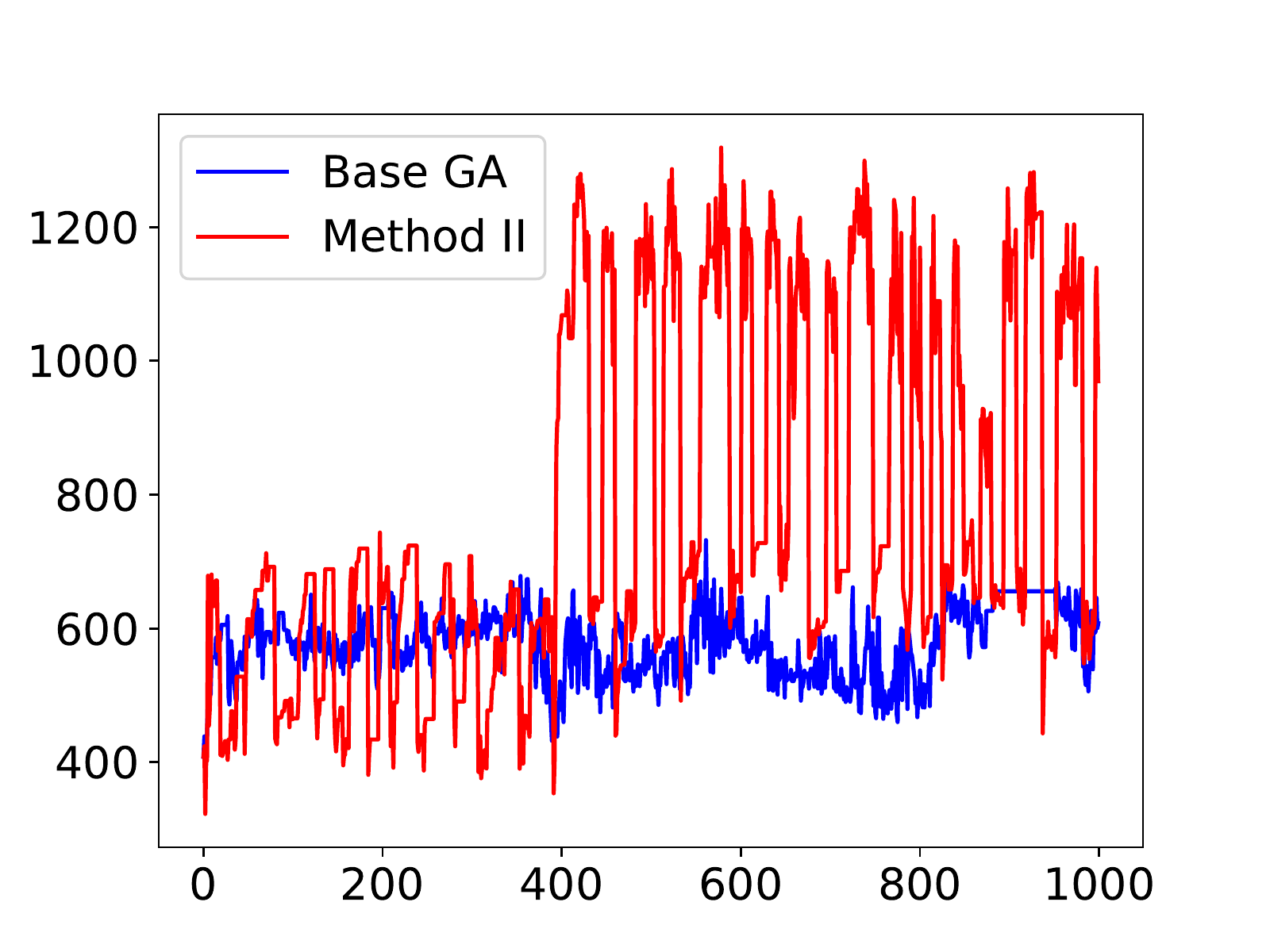}\\
	\end{tabular}
	\caption{Base GA and Method II learning progress. Mean denotes population mean game score over generations in training, high denotes score of top-performaing individual over generations in training, and validation denotes the mean score of the best-generalizing individual to 30 differently-seeded environments. In each generation, the best individual in validation is designated as the elite. In 3 out of 4 games, validation scores reach a higher maximum. Whereas the Base GA seemingly failed to escape a local optima, Method II was particularly effective for improving  performance in \textsc{Space Invaders}.}
	\label{fig:method2-basega}
\end{figure*}

We then evaluated Method 2 by comparing it to the Base GA. These experiments were run using similar hyperparameters to related work \cite{such2017deep} --- (see Table \ref{table:method2params}). Method II training progress is visualized by Figure \ref{fig:method2-basega} and testing evaluation of Method II is summarized by Table \ref{table:method2results}. In testing, Method II yielded improved results over the Base GA in two out of four games and no significant change in two out of four games. We also compared Method II testing scores to those reported in \cite{mnih2015human} for Deep Q-learning --- see Table \ref{table:method2vsDQN}. Method II outperforms DQN methods in one game, and is outperformed by DQN methods in two games. These mixed results are consistent with previous work \cite{such2017deep}.

\begin{table}
	\begin{center}
		
		\begin{tabular}{lrrrr}
			\hline 
			& \multicolumn{2}{c}{ Mean} &   \multicolumn{2}{c}{ St.\ Dev.}    \\ 
			Game &  Random &  Method II   &  Random  &  Method II    \\
			\hline 
			\textsc{MsPacman} & 3377 & \textbf{3790} & 661 & 322   \\
			\hline 
		\end{tabular} 
	\end{center}
	\caption{Comparison of Method II (novelty-based population resampling) to random population-resampling over 30 episodes not used in training or validation. In \textsc{MsPacman}, Method II yielded better mean game scores in testing than random population resampling with p $< 0.05$ in a two-tailed t-test.}
	\label{table:randomresults}
\end{table}

\begin{table}
	\begin{center}
		
		\begin{tabular}{lrrrr}
			\hline 
			& \multicolumn{2}{c}{ Mean} &   \multicolumn{2}{c}{ St.\ Dev.}     \\ 
			Game &  Base GA &  Method II  &  Base GA  &  Method II    \\
			\hline 
			\textsc{Assault} & 1219 & 1007 & 676 & 413  \\
			\textsc{Asteroids} & 1263 & 1476 & 590 & 640   \\
			\textsc{MsPacman} & 3385 & \textbf{3700} & 633 & 209  \\
			\textsc{Space Invaders} & 615 & \textbf{1211} & 323 & 244   \\
			\hline 
		\end{tabular} 
	\end{center}
	\caption{Comparison of Base GA and Method II testing results over 30 episodes not used in training or validation. Means and standard deviations are measured in game score units. Bolded means denote significantly better testing performance (p < $0.05$ in a two-tailed t-test). Method II improves learning in 2 out of 4 games over the Base GA.}
	\label{table:method2results}
\end{table}

\begin{table}
	\begin{center}
		\begin{tabular}{lrrrr}
			\hline 
			& \multicolumn{2}{c}{ Mean} &   \multicolumn{2}{c}{ St.\ Dev.}    \\ 
			Game &  DQN &  Method II  &  DQN  &  Method II    \\
			\hline 
			\textsc{Assault} & \textbf{3359} & 1007 & 775 & 413   \\
			\textsc{Asteroids} & 1629 & 1476 & 542 & 640   \\
			\textsc{MsPacman} & 2311 & \textbf{3700} & 525 & 209  \\
			\textsc{Space Invaders} & \textbf{1976} & 1211 & 893 & 244   \\
			\hline 
		\end{tabular} 
	\end{center}
	\caption{Comparison of DQN and Method II using testing scores over 30 randomly-seeded episodes reported in \cite{mnih2015human}.  Means and standard deviations are measured in game score units. Means and standard deviations are measured in game score units. Bolded means denote significantly better testing performance (p < $0.05$ in a two-tailed t-test). Method II outperforms DQN in one game, performs similarly to DQN in one game, and is outperformed by DQN in two games. These mixed results are consistent with previous comparisons between gradient-based and gradient-free learning methods \cite{such2017deep}.}
	\label{table:method2vsDQN}
\end{table}

\section{Discussion}\label{section:discussion}

The results presented in this paper support recent work showing that GAs are effective at training deep neural networks for RL. We took advantage of this to explore whether the behaviour of agents could be effectively used as selection pressure in an evolutionary search for RL policies. While our implementation of novelty search based on Levenshtein distance was not as effective as the Base GA, we found that it produced potentially useful and informative policies. In particular, we found that novelty search over Levenshtein distances is not equivalent to a longevity search, and that the policies it produces may be more defensive than than those produced by typical reward optimization. 

The combination of reward signal and novelty scores in Method II resulted in a net improvement in testing scores over the Base GA in the four games tested. During \textsc{Space Invaders} training it is particularly evident that, while the Base GA was showing signs of stagnation or convergence in validation performance, Method II effectively reoriented the search. 

Method II yielded improved policies for MsPacman over both the Base GA and DQN method. On closer inspection, however, it is clear that all of the compared policies suffer from limited complexity. In no cases did we observe a successful strategy shift from escape to pursuit upon consumption of a pill. The lack of this emergent behaviour in any of the results we considered, in addition to sub-human performance in the current state-of-the-art based on a modified DQN architecture \cite{hessel2018rainbow}, leads us to suspect that the DQN architecture combined with reward-signal optimization is not well-suited for effectively learning situational policies or \textit{discrete mode switching}. In response, we think that emerging methods such as Differentiable Inductive Logic Programming \cite{evans2018learning}, a learning framework that enables logical rules to be inferred from large-scale data using neural networks, and a new wave of automated network architecture construction algorithms could be especially useful.

More broadly, the production and storage of policies with varying behaviours, including defensiveness, could have many applications in real-world control problems. In autonomous transport, for example, it could be desirable to evaluate potential policies with a wide range of behaviours in order to select the safest. Methods based on novelty search, like the ones introduced in this paper, could be used to purposefully learn diverse strategies for achieving the same goals. This concept has recently been shown to be effective in learning frameworks based on a wide variety of methods --- see \cite{goexplore} and \cite{savinov2018episodic}, each of which use environment observations to help instil novelty, in addition to \cite{mouret2015illuminating} and \cite{pugh2016quality}. Methods that already implement observationally-based storage and comparison methods could benefit from the relatively low-cost inclusion of action sequences and string edit metric distances to diversify learned policies.

\section{Future Work}

The evolutionary algorithms community has developed and applied many methods for evolving network architectures and related structures. NEAT \cite{stanley2002evolving} and HyperNEAT \cite{stanleyhypercube} are very popular methods for simultaneously evolving network architectures and weights, and Cartesian Genetic Programming \cite{miller2011cartesian} is a related method that uses more general basis functions than are typically used in neural networks. All of these methods have been successfully applied in RL problems. 

In future work, we will extend the methods detailed in this paper to include automated network architecture search. A method inspired by NEAT and that uses a compact algebraic approach to modular network representation \cite{jackson2017algebraic} is currently in development. Given the scale at which Such \etal.'s method enables GAs to train deep neural networks, we are optimistic that both existing and forthcoming methods for topology- and weight- evolving neural networks (TWEANNs) will be effective tools for solving increasingly complex problems in RL.

We are particularly eager to develop tools that combine the open-endedness of evolutionary algorithms with the reliability and robustness of functional modules, which could range from simple logical operators to convolutional network layers and beyond. Methods for searching the complex search space of deep neural network architectures and hyperparameters have recently been developed for gradient-based learning \cite{negrinho2017deeparchitect}. And though similar methods like HyperNEAT are certainly able to learn high-quality RL policies \cite{:2019aa}, we think a method that combines recent advances in both evolutionary algorithms and gradient-based deep reinforcement learning could be even more effective.

%\clearpage

\bibliographystyle{ACM-Reference-Format}
\bibliography{GECCO_2019_bibliography} 

\clearpage
\section{Appendix}

\begin{algorithm}
	\caption{Base GA}\label{alg:baseGA}
	\begin{algorithmic}[0]
		\State \textbf{Input:} mutation function $\psi$, population size $N$, number of generations $G$, truncation size $T$, individual initializer $\phi$, individual decoder $\gamma$, fitness function $F$, training episodes $E_{t}$, validation episodes $E_{v}$, deterministic uniform PRNG $U$.
		\State $\text{\emph{population}} \gets \text{[]}$
		\For{$i = 1, 2, \dots, N$} 
		% \State $\text{population} \gets \text{population} + [\phi(U(0,2^{32}-1))$]
		\State \textbf{Append} $\phi(U(0,2^{32}-1))$ to \emph{population}
		\EndFor
		\For{$g = 1, 2, \dots, G$} 
		\State \emph{policies} $\gets$ $\text{map}(\gamma, \text{\emph{population}})$
		\State \emph{trainingResults} $\gets$ $F(E_{t}$, \emph{policies})
		\State \textbf{Sort} \emph{trainingResults} by game score
		\State \emph{eliteCandidates} $\gets$ $10$ best in \emph{trainingResults}
		\State \emph{validationResults} $\gets$ $F(E_{v}$, \emph{eliteCandidates})
		\State \textbf{Sort} \emph{validationResults} by game score
		\State \emph{elite} $\gets$ 1 best in \emph{validationResults}
		\State \textbf{Save} \emph{elite} to disk
		\State \emph{parents} $\gets$ $T$ best in \emph{trainingResults}
		
		\If {$g < G - 1$}
		\State \emph{newPopulation} $\gets$ [\emph{elite}]
		\For{$p = 1, 2, \dots N-1$}
		\State \emph{parent} $\gets \text{\emph{parents}}[U(0,T-1)]$
		\State \textbf{Append} $\psi$(\emph{parent}) to \emph{newPopulation}
		% \State newPopulation $\gets$ newPopulation + [$\psi$(parent)]
		\EndFor
		\State \emph{population} $\gets$ \emph{newPopulation}
		\EndIf
		\EndFor
	\end{algorithmic}
\end{algorithm}

\begin{algorithm}
	\caption{Method I - Novelty Search}\label{alg:noveltyGA}
	\begin{algorithmic}[0]
		\State \textbf{Input:} mutation function $\psi$, population size $N$, number of generations $G$, truncation size $T$, individual initializer $\phi$, individual decoder $\gamma$, fitness function $F$, training episodes $E_{t}$, validation episodes $E_{v}$, deterministic uniform PRNG $U$, archive insertion probability $p$, novelty function $\eta$.
		\State $\text{\emph{population}} \gets \text{[]}$
		\For{$i = 1, 2, \dots, N$} 
		% \State $\text{population} \gets \text{population} + [\phi(U(0,2^{32}-1))$]
		\State \textbf{Append} $\phi(U(0,2^{32}-1))$ to \emph{population}
		\EndFor
		\State $A \gets$ [] 
		\For{$g = 1, 2, \dots, G$} 
		\State \emph{policies} $\gets$ $\text{map}(\gamma, \text{\emph{population}})$
		\State \emph{trainingResults} $\gets$ $F(E_{t}$, \emph{policies})
		\For{$(ind, gameScore, BC)$ in \emph{trainingResults}}
		\State \textbf{Append} $BC$ to $A$ with probability $p$
		\EndFor
		\State \emph{nScores} $\gets$ map($\eta(A), \emph{trainingResults})$
		\State \textbf{Sort} \emph{trainingResults} by novelty score
		\State \emph{eliteCandidates} $\gets$ $10$ best in \emph{trainingResults}
		\State validationResults $\gets$ $F(E_{v}$, \emph{eliteCandidates})
		\State \textbf{Sort} \emph{validationResults} by game score
		\State \emph{elite} $\gets$ 1 best in \emph{validationResults}
		\State \textbf{Save} \emph{elite} to disk
		\State \emph{parents} $\gets$ $T$ most novel in \emph{trainingResults}
		
		\If {$g < G - 1$}
		\State \emph{newPopulation} $\gets$ [\emph{elite}]
		\For{$p = 1, 2, \dots N-1$}
		\State \emph{parent} $\gets \text{\emph{parents}}[U(0,T-1)]$
		\State \textbf{Append} $\psi$(\emph{parent}) to \emph{newPopulation}
		% \State newPopulation $\gets$ newPopulation + [$\psi$(parent)]
		\EndFor
		\State \emph{population} $\gets$ \emph{newPopulation}
		\EndIf
		\EndFor
	\end{algorithmic}
\end{algorithm}

\begin{algorithm}
	\caption{Method II - Stagnation Detection and Population Resampling}\label{alg:methodIIGA}
	\begin{algorithmic}[0]
		\State \textbf{Input:} mutation function $\psi$, population size $N$, number of generations $G$, truncation size $T$, individual initializer $\phi$, individual decoder $\gamma$, fitness function $F$, training episodes $E_{t}$, validation episodes $E_{v}$, deterministic uniform PRNG $U$, archive insertion probability $p$, novelty function $\eta$, number of improvement generations $IG$
		\State $\text{\emph{population}} \gets \text{[]}$
		\For{$i = 1, 2, \dots, N$} 
		% \State $\text{population} \gets \text{population} + [\phi(U(0,2^{32}-1))$]
		\State \textbf{Append} $\phi(U(0,2^{32}-1))$ to \emph{population}
		\EndFor
		\State \emph{vScores} $\gets$ []
		\For{$g = 1, 2, \dots, G$} 
		\State \emph{policies} $\gets$ $\text{map}(\gamma, \text{\emph{population}})$
		\State \emph{trainingResults} $\gets$ $F(E_{t}$, \emph{policies})
		\For{$(ind, gameScore, BC)$ in \emph{trainingResults}}
		\State \textbf{Append} $BC$ to $A$ with probability $p$
		\EndFor
		\State \textbf{Sort} \emph{trainingResults} by game score
		\State \emph{eliteCandidates} $\gets$ $10$ best in \emph{trainingResults}
		\State \emph{validationResults} $\gets$ $F(E_{v}$, \emph{eliteCandidates})
		\State \textbf{Sort} \emph{validationResults} by game score
		\State \emph{elite} $\gets$ 1 best in \emph{validationResults}
		\State \textbf{Save} \emph{elite} to disk
		\State \textbf{Append} \emph{elite} validation score to \emph{vScores}
		\State \emph{parents} $\gets$ $T$ best in \emph{trainingResults}
		\If {$vScores.length \geq IG$}
		\State \emph{progress} $\gets$ []
		\For{$i=g-IG+1, g-IG+2, ... g$}
		\State \textbf{Append} \emph{vScores}$[i]$ - \emph{vScores}$[g-IG]$ to \emph{progress}
		\EndFor
		\If{$\forall x$ in \emph{progress}, $x <= 0$}
		\State \emph{noveltyResults} $\gets$ map($\eta(\emph{trainingResults}), A)$
		\State \textbf{Sort} \emph{noveltyResults} by novelty score
		\State \emph{parents} $\gets$ $T$ most novel in \emph{noveltyResults}
		\State \emph{vScores} $\gets$ []
		\EndIf
		\EndIf
		\If {$g < G - 1$}
		\State \emph{newPopulation} $\gets$ [\emph{elite}]
		\For{$p = 1, 2, \dots N-1$}
		\State \emph{parent} $\gets \text{\emph{parents}}[U(0,T-1)]$
		\State \textbf{Append} $\psi$(\emph{parent}) to \emph{newPopulation}
		% \State newPopulation $\gets$ newPopulation + [$\psi$(parent)]
		\EndFor
		\State \emph{population} $\gets$ \emph{newPopulation}
		\EndIf
		\EndFor
	\end{algorithmic}
\end{algorithm}

\end{document}